\documentclass{article}

\PassOptionsToPackage{sort&compress}{natbib}
\usepackage[preprint]{corl_2026} % arXiv preprint: real authors, no line numbers, no CoRL footnote

\usepackage{iftex}
\ifXeTeX
  \usepackage{fontspec}
\fi
\usepackage{amsmath,amssymb}
\usepackage{graphicx}
\graphicspath{{./}}
\usepackage{booktabs}
\usepackage{multirow}
\usepackage[table]{xcolor}
\usepackage{subcaption}
\usepackage{enumitem}
\usepackage{wrapfig}
\usepackage{float}
\usepackage{tabularx}
\usepackage{longtable}
\usepackage{url}

\newcommand{\ours}{HumanEgo}
\newcommand{\geo}{\textsc{ICT}}

\newcommand{\eg}{\emph{e.g.}}

\newcommand{\SE}{$\mathrm{SE}(3)$}

\newcommand{\R}{\mathbb{R}}

\title{\ours{}: Zero-Shot Robot Learning \\ from Minutes of Human Egocentric Videos} % Sample-Efficient Cross-Embodimen Embodiment-Agnostic

\author{
  \textbf{Zhi (Leo) Wang} \quad \textbf{Botao He} \quad \textbf{Kelin Yu} \quad \textbf{Seungjae Lee} \\
  \textbf{Ruohan Gao} \quad \textbf{Furong Huang} \quad \textbf{Yiannis Aloimonos} \\[3pt]
  {\normalfont University of Maryland} \\[3pt]
  {\normalfont Project page: \url{https://humanego-ai.github.io/}} \\[1pt]
  {\normalfont Code: \url{https://github.com/TX-Leo/HumanEgo}}
}

\makeatletter
\renewcommand{\@maketitle}{%
  \vbox{%
    \hsize\textwidth
    \linewidth\hsize
    \vskip 0.05in
    \centering
    {\LARGE\bf \@title\par}
    \def\And{%
      \end{tabular}\hfil\linebreak[0]\hfil%
      \begin{tabular}[t]{c}\bf\rule{\z@}{8\p@}\ignorespaces%
    }%
    \def\AND{%
      \end{tabular}\hfil\linebreak[4]\hfil%
      \begin{tabular}[t]{c}\bf\rule{\z@}{8\p@}\ignorespaces%
    }%
    \begin{tabular}[t]{c}\bf\rule{\z@}{8\p@}\@author\end{tabular}%
    \vskip 0.2in \@minus 0.1in
  }%
}
\makeatother

\usepackage{titlesec}
\titlespacing*{\section}      {0pt}{2pt plus 1pt minus 1pt}{0.5pt plus 0.3pt}
\titlespacing*{\subsection}   {0pt}{1.5pt plus 1pt minus 1pt}{0.4pt plus 0.3pt}
\titlespacing*{\subsubsection}{0pt}{2pt plus 1pt minus 1pt}{0.6pt plus 0.5pt}
\titlespacing*{\paragraph}    {0pt}{2pt plus 1pt minus 1pt}{0.8em}

\begin{document}
\maketitle

\vspace{-10pt}
\begin{figure}[H]
  \centering
  \vspace{-5pt}
  \includegraphics[width=\textwidth]{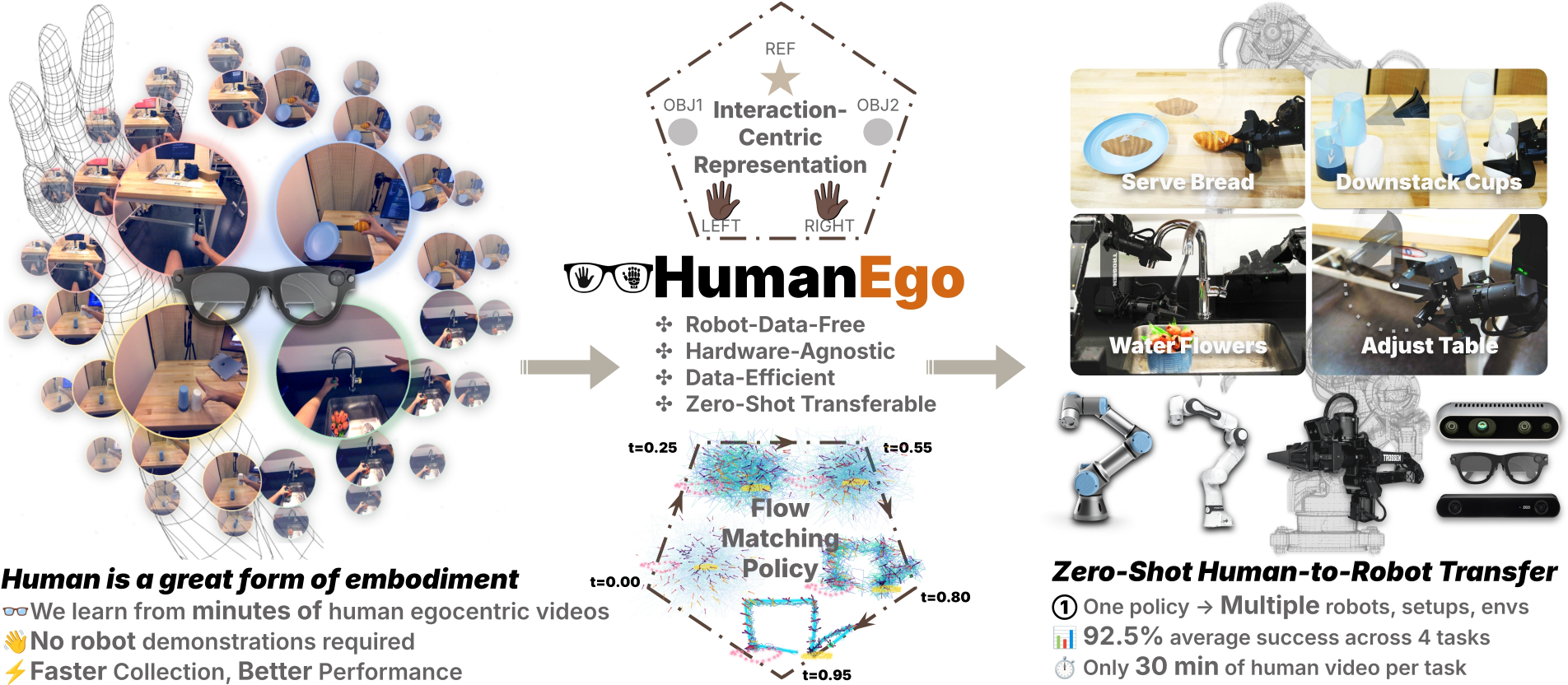}
  \vspace{-8pt}
  \caption{\textbf{\ours{} learns robot policy from human egocentric videos.}
  A human wears Aria glasses and collects demonstrations~(\textbf{left}); the egocentric videos are converted into an interaction-centric representation and used to train a flow matching policy~(\textbf{middle}); the policy transfers \emph{zero-shot} to the robot---free of environment, setup, or embodiment~(\textbf{right}).}
  \label{fig:teaser}
  \vspace{-8pt}
\end{figure}

\begin{abstract}
Human egocentric video captures rich manipulation demonstrations without any robot hardware, yet transferring these skills to robots remains challenging due to the embodiment gap between human and robot in both visual appearance and kinematics.
We present \textbf{\ours{}}, a framework that bridges the embodiment gap by lifting each human demonstration to an entity-level representation of \emph{hand--object interaction}, and training a \emph{flow matching} policy with \emph{dense auxiliary objectives} that amplify supervision from every trajectory. \ours{} is robot-data-free, hardware-agnostic, data-efficient, and zero-shot human-to-robot transferable.
With only 30~minutes of human videos per task, \ours{} achieves \textbf{92.5}\% average success across four real-world tasks (\textbf{75}\% with just 15~minutes), outperforms matched-time robot teleoperation by \textbf{41}~percentage points, and robustly transfers zero-shot across novel robots, cameras, and environments.
We release \ours{} as an easy-to-use, open-source framework for learning robot
policies directly from human data: \url{https://github.com/TX-Leo/HumanEgo}.
\end{abstract}
\vspace{-6pt}

%==============================================================================

\section{Introduction}
\label{sec:intro}

% ¶1 — Motivation + Landscape + Our Goal
State-of-the-art manipulation policies require hundreds to thousands
of task-specific robot demonstrations~\citep{brohan2023rt2,aloha2024aloha2,chi2023diffusionpolicy,khazatsky2024droid,kim2024openvla,black2025pi},
which are costly, time-consuming, and inconvenient to collect.
Human egocentric video offers a much cheaper and more accessible
alternative: with a head-mounted camera~\citep{engel2023aria}, a
single person can collect task demonstrations anywhere, in minutes.
But how should we leverage this data?
Existing approaches fall into two paradigms, each with significant
limitations.
\emph{Co-training} methods~\citep{kareer2024egomimic,punamiya2025egobridge,liu2025immimic,qiu2025humanoid}
supplement robot data with human video, but still require substantial robot
demonstrations for every new task---reducing, rather than eliminating, the
data burden.
\emph{Large-scale pretraining}
approaches~\citep{yang2025egovla,zheng2026egoscale,punamiya2026egoverse,hoque2026egodex} learn from massive
egocentric corpora, but demand enormous compute and still require
robot-specific post-training to produce deployable policies.
We pursue a more direct goal:
\textbf{learning deployable manipulation policies from only minutes of
human egocentric demonstrations---without any robot data or
internet-scale pretraining.}

% ¶2 — Representation Challenge
Achieving this goal exposes two fundamental challenges.
\textbf{(1)~The representation challenge: bridging the embodiment gap.}
Humans and robots differ in both \emph{visual appearance} and
\emph{kinematics}, and these gaps demand distinct solutions.
On the visual side, retargeting-based
methods~\citep{lepert2025phantom,lepert2025masquerade,dessalene2025embodiswap} synthesize robot-like
imagery from human video but are brittle to morphological and viewpoint
differences;
point-tracking approaches~\citep{bharadhwaj2024track2act,haldar2025pointpolicy,ze2024egozero}
extract sparse geometric features but discard the rich visual context
surrounding interactions.
On the kinematic side, hierarchical
methods~\citep{wang2023mimicplay,wang2024h2r,xu2023xskill} separate high-level plans
from low-level execution but still require robot data for the low-level
controller;
object-centric approaches~\citep{xu2024im2flow2act,jain2024vid2robot,atm}
track only the manipulated object, losing critical information about
\emph{how} the hand approaches, grasps, and releases it.
We argue that neither hand nor object alone defines a
skill---what matters is their \emph{interaction}.
This is the key representational claim behind \ours{}: robots
should not imitate the human body, but recover the
task-relevant \emph{interaction geometry} that transfers across
bodies.

% ¶3 — Learning Challenge
\textbf{(2)~The learning challenge: learning from minimal data.}
Although raw human video is abundant online, clean clips with
precise action labels remain scarce, making data-efficient learning
from minutes of per-task videos critical. This regime introduces
two distinct challenges: \emph{multi-modality} and \emph{signal
sparsity}.
For the multi-modality challenge, the same task admits many valid
strategies. Diffusion-based methods~\citep{chi2023diffusionpolicy}
capture this distribution but need many denoising steps and are slow
at inference; faster alternatives~\citep{zhao2023act} are less
expressive.
For the signal-sparsity challenge, each trajectory carries rich
signal beyond the hand action---object motion, visual traces,
hand--object state---yet prior work taps only a fraction: single
auxiliary targets such as visual foresight~\citep{xu2024im2flow2act,yu2025genflowrl,li2025novaflow}
or 2D tracks~\citep{bharadhwaj2024track2act,atm,patel2025rigvid}, or upstream
pretraining corpora~\citep{yang2025egovla,zheng2026egoscale,yu2024mimictouch}.
We argue that a fast generative policy paired with multi-type dense
supervision is the key to data-efficient learning from minutes of
human egocentric videos.
In other words, the goal is to squeeze many forms of supervision
from human video, so small curated demonstrations
can punch above their size.

% ¶4 — Our Approach
We present \ours{}~(Fig.~\ref{fig:teaser}), addressing each gap with a
targeted design.
For the \emph{visual gap}, we inpaint the human arm from each
egocentric frame and render a virtual gripper with tracked object
keypoints in its place, producing an embodiment-agnostic visual
observation.
For the \emph{kinematic gap}, we encode every hand and object as an
Interaction-Centric Token~(ICT), producing a compact,
embodiment-invariant spatial observation whose hand--entity blocks are
viewpoint-invariant by construction.
For \emph{multimodality}, we adopt a flow matching~\citep{lipman2023flow}
policy, producing expressive multi-modal actions at fast inference.
For \emph{signal sparsity}, we design three dense auxiliary
objectives: 2D trace, object motion, and latent consistency.
Together they produce multi-type dense supervision from each
trajectory's scene dynamics, boosting learning from few
demonstrations. Our contributions:
\begin{itemize}[itemsep=1pt, topsep=2pt, leftmargin=15pt, parsep=0pt]
  \item \textbf{\ours{}}, a robot-data-free, hardware-agnostic, and
    data-efficient pipeline that learns robot manipulation policies from
    minutes of raw human egocentric videos---powered by a transferable
    interaction representation and a flow matching policy with
    dense auxiliary objectives.
  \item \textbf{Interaction-Centric Tokens~(\geo{})}, a compact
    entity-level representation of hand--object interaction that is
    embodiment-invariant, with hand--entity blocks that are exactly
    viewpoint-invariant.
  \item \textbf{Robust zero-shot human-to-robot transfer.}
    Trained on 30~minutes of human video per task, \ours{} reaches
    \textbf{92.5}\% success across 4 real-world tasks and \textbf{75}\%
    at half that budget. At matched collection time, it surpasses
    robot teleoperation by \textbf{41}~percentage points. The learned policy also
    deploys zero-shot to novel robot embodiments, camera setups,
    lighting, backgrounds, and object instances, without any
    retraining or fine-tuning. This suggests that human video is not
    merely a cheap substitute but a \emph{scalable} and
    \emph{potentially superior} data source for policy learning.
\end{itemize}

\section{Related Work}
\label{sec:related}

%===============================================================================
% Architecture — floats to top of Page 3
%===============================================================================
\begin{figure}[!t]
    \centering
    \includegraphics[width=0.79\textwidth]{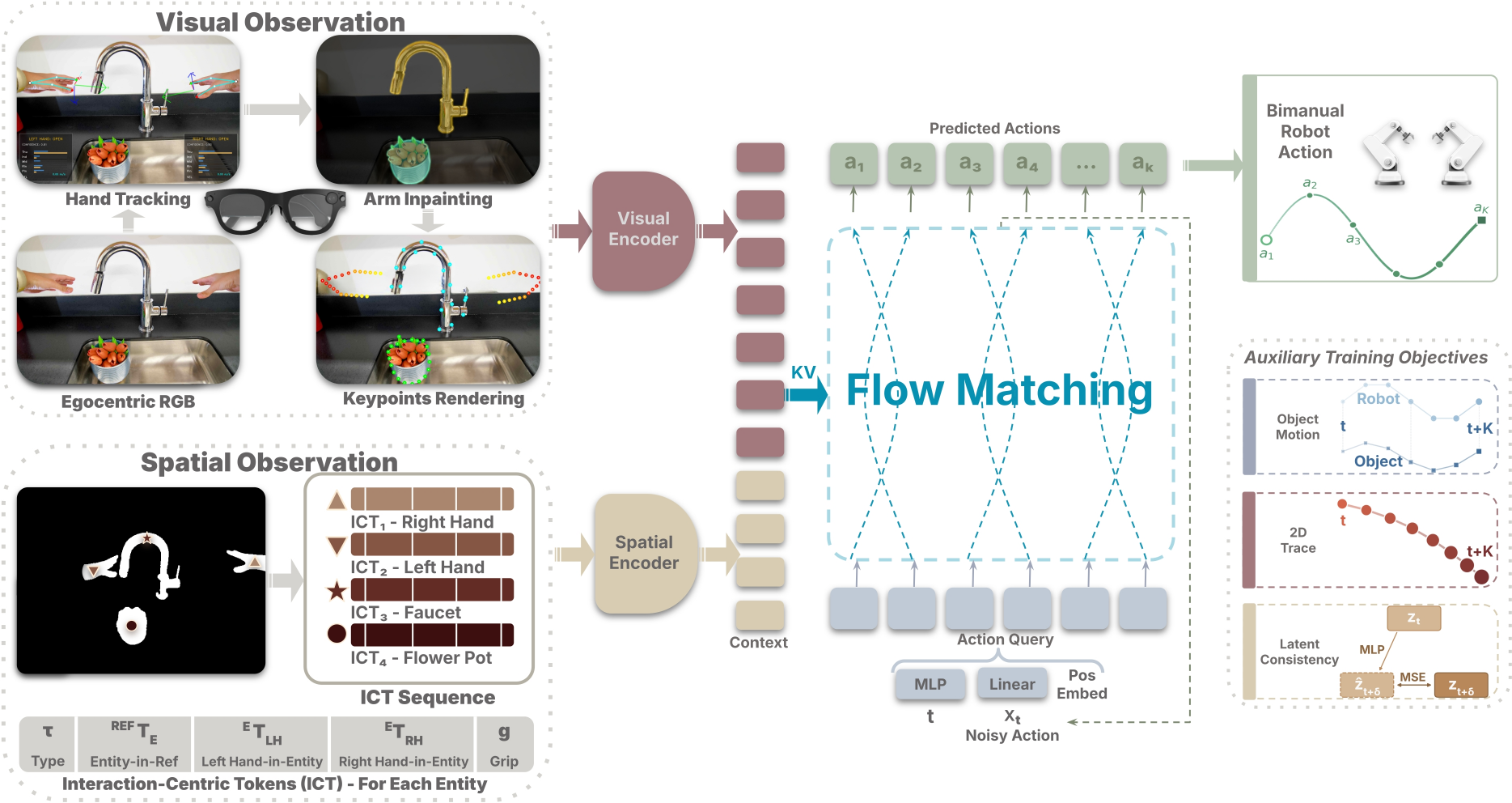}
    \caption{\textbf{System overview of \ours{}.} Arm inpainting and visual keypoints bridge the visual gap;
    Interaction-Centric Tokens encode spatial relationships among all entities;
    a flow matching policy with dense auxiliary objectives learns bimanual robot actions from minutes-scale human data.}
    \label{fig:system_overview}
\end{figure}

Large-scale egocentric and hand--object interaction
datasets~\citep{grauman2022ego4d,damen2022epickitchens100,hoque2026egodex,banerjee2024hot3d,liu2022hoi4d,liu2024taco,wang2023holoassist,punamiya2026egoverse}
underpin learning manipulation from human video. Generalist policies and
world models~\citep{yang2025egovla,zheng2026egoscale,zhang2026unidex,lee2025tracegen,yuan2024general}
learn embodiment-agnostic representations from massive corpora, but demand
heavy compute and per-task robot post-training;
co-training~\citep{kareer2024egomimic,punamiya2025egobridge,jain2024vid2robot,zhu2025emma,liu2025immimic,kareer2025human2robot,qiu2025humanoid}
and visual retargeting~\citep{lepert2025phantom,wang2024h2r,lepert2025masquerade,dessalene2025embodiswap}
still need robot demonstrations or renderings brittle to morphology and
viewpoint; and hierarchical
methods~\citep{wang2023mimicplay,xu2023xskill,kim2025uniskill} delegate
low-level control to a robot-trained controller. Zero-shot methods avoid
robot data entirely and differ in \emph{what} they represent:
points~\citep{ze2024egozero,haldar2025pointpolicy,guzey2025aina,singh2025afford2act},
object pose or motion~\citep{hsu2024spot,zou2026activeglasses,patel2025rigvid,yu2025genflowrl,yin2025motionfield},
or image goals~\citep{bharadhwaj2024track2act,he2024zeromimic}, with further
directions explored in~\citep{park2025demodiffusion,xu2024im2flow2act,shah2025mimicdroid,li2025novaflow,chen2026videomanip,shi2025points2reward,planner,watch}.
All encode the hand \emph{or} the object, rarely their \emph{interaction}---the
very signal that defines manipulation. \ours{} encodes it explicitly, achieving
zero-shot transfer from minutes of human video without robot data or
large-scale pretraining. Appendix~\ref{app:related} expands this discussion.

% \paragraph{Data-efficient imitation learning.}
% Even when a suitable representation is available, learning robust
% policies from a handful of demonstrations remains a fundamental
% bottleneck.
% The field has pursued two complementary strategies.
% The first is \emph{data augmentation}: manufacturing additional training
% signal through image transformations and action
% perturbation~\citep{chi2023diffusionpolicy,zhao2023act}, or leveraging
% rendering engines that synthesize diverse visual
% variations~\citep{lepert2025phantom,wang2024h2r}.
% Video world models~\citep{du2025dreamgen} push this idea to its logical
% extreme by hallucinating entire rollouts from a few seed demonstrations.
% The second strategy introduces \emph{auxiliary objectives} that force the
% network to extract richer structure from the same limited
% data---\eg{}, auxiliary visual cues for spatial
% awareness~\citep{yang2025aimbot} or flow prediction as a self-supervised
% proxy~\citep{xu2024im2flow2act}.
% \ours{} unifies both: we augment the training set for coverage and
% equip a flow matching policy~\citep{lipman2023flow} with dense auxiliary
% objectives---2D trace prediction, object dynamics forecasting, and
% temporal consistency---that provide per-timestep supervision well beyond
% the primary action loss, squeezing maximum learning signal from every
% human demonstration.

%===============================================================================

\section{\ours{}}
\label{sec:method}

\ours{} turns human egocentric video into a deployable bimanual
policy in five stages (Fig.~\ref{fig:system_overview}).
A demonstrator wearing Aria glasses records the task
(Sec.~\ref{sec:data_collection}); we close the \emph{embodiment gap}
by inpainting the human arm and rendering a virtual gripper
(Sec.~\ref{sec:visual}) and by encoding every entity's pose relative
to other task entities into Interaction-Centric Tokens
(Sec.~\ref{sec:spatial}). A flow matching policy with
three auxiliary objectives generates multi-modal bimanual actions
(Sec.~\ref{sec:policy}), and the same policy drives a real robot by
swapping only the modules that produce its inputs
(Sec.~\ref{sec:deployment}).

\subsection{Egocentric Data Collection}
\label{sec:data_collection}

A human demonstrator wearing Aria Gen1
glasses~\citep{engel2023aria} performs the target task in any
convenient environment---regardless of table height, lighting, or
background, and without specialized workspace or calibration.
Demonstrations are recorded at 30\,Hz.
Aria glasses are particularly well suited for learning from human video:
their Machine Perception Services (MPS) provide high-quality 6-DoF SLAM
tracking, calibrated 3D hand pose estimation, and synchronized egocentric
RGB streams---all from a single lightweight wearable device.

\subsection{Visual Observation Preprocessing}
\label{sec:visual}

We transform the undistorted egocentric frames into
embodiment-agnostic RGB observations in two steps.
First, we segment the human hand and arm with SAM2~\citep{ravi2024sam2} and remove them via
LaMa inpainting~\citep{suvorov2022lama}, eliminating the visual
embodiment gap.
Second, we render a virtual gripper and the tracked object keypoints
into the inpainted image---both derived from the spatial observation
(Sec.~\ref{sec:spatial})---implicitly encoding 6D pose information as
visual cues.
This lightweight procedure bridges the visual embodiment gap without
expensive domain adaptation or image translation.

\subsection{Spatial Observation Preprocessing}
\label{sec:spatial}

We build our explicit entity-level spatial observation: treating every
object and both hands as an \emph{entity}, we track the hands and
objects to recover each entity's 6-DoF pose, then encode their
relative relations into Interaction-Centric Tokens. We detail these
three steps below:

\textbf{Hand tracking and motion optimization.}
We start from the 3D hand keypoints produced by Aria
MPS~\citep{engel2023aria}, lift them to the world frame via SLAM, and
smooth them with Savitzky--Golay on positions and an exponential
moving average (EMA) on rotations.
We then treat the thumb--index pair as a virtual parallel-jaw gripper, extracting an $\mathrm{SE}(3)$
end-effector pose $T_{\text{ee}}$ and a scalar grasp $g$.
Position is the fingertip midpoint; orientation is a Gram--Schmidt frame
built on the metacarpophalangeal (MCP) joints rather than the fingertips,
which avoids the degeneracy when fingertips converge during pinch grasps;
and the grasp scalar $g\in[0,1]$ normalizes the thumb--index fingertip
distance, binarized at deployment (Appendix~\ref{app:hand2gripper}).

\textbf{Object tracking and pose estimation.}
We detect each object with text-prompted Grounding
DINO~\citep{liu2024groundingdino}, segment it with
SAM2~\citep{ravi2024sam2}, and sample contour keypoints from the mask.
We track these 2D keypoints $\mathbf{u}_n$ across the video with
CoTracker3~\citep{karaev2024cotracker} and lift them to 3D via
$\mathbf{p}_n = \mathrm{Triangulate}(\mathbf{u}_n,\, K,\, T_{\text{SLAM}})$,
using camera intrinsics $K$ and the per-frame Aria SLAM pose
$T_{\text{SLAM}}$.
We take the centroid of the $N$ tracked points as the object
position to cancel per-point triangulation noise,
$\mathbf{p}_{\text{obj}} = \tfrac{1}{N}\sum_{n=1}^{N} \mathbf{p}_n$,
and estimate orientation $R_{\text{obj}}$ with
Orient-Anything V2~\citep{wu2025orientanything}.
During grasping the object is occluded by the hand, so we apply
\emph{kinematic latching}---rigidly tying the object pose to the hand
from the grasp onset $t_0$:
$T_{\text{obj}}^{t} = T_{\text{hand}}^{t} \cdot
(T_{\text{hand}}^{t_0})^{-1}\, T_{\text{obj}}^{t_0}$.
The latch replaces an estimate that occlusion has made unreliable with
one propagated by the hand, and assumes the object does not move
relative to the hand once grasped.

\textbf{Entity Spatial Encoding via Interaction-Centric Tokens~(\geo{}).}
\label{sec:tokens}
We encode each entity's 6-DoF pose into an \geo{}, capturing both
its pose in a shared reference frame and its spatial relation to
both hands.
For each entity $k{=}1,\ldots,N$, the token $\geo{}_k \in \R^{29}$ is:
\begin{equation}
    \geo{}_k = [\underbrace{\tau}_{1} \;\|\;
           \underbrace{{}^{\mathrm{REF}}\!T_{E}}_{9} \;\|\;
           \underbrace{{}^{E}\!T_{LH}}_{9} \;\|\;
           \underbrace{{}^{E}\!T_{RH}}_{9} \;\|\;
           \underbrace{g}_{1}],
    \label{eq:ict}
\end{equation}
where $\tau$ is the entity type (hand or object);
${}^{\mathrm{REF}}\!T_{E}$ is entity $k$'s pose in a reference frame
$\mathrm{REF}$ that is held fixed for the episode---the camera pose at the
onset of manipulation during training, where the head-mounted camera
moves, and the static camera itself at deployment
(Sec.~\ref{sec:deployment});
${}^{E}\!T_{LH}$ and ${}^{E}\!T_{RH}$ are the left-hand ($LH$) and
right-hand ($RH$) poses expressed in entity $k$'s local frame $E$;
and $g$ is the grasp state (binarized finger distance for hands;
a sentinel for objects).
We flatten each SE(3) transform to a 9D vector by concatenating the
normalized translation with a 6D rotation
representation~\citep{zhou2019continuity}, and derive every quantity
from off-the-shelf perception without ground-truth labels.
Unlike prior methods using global point clouds or absolute
coordinates~\citep{ze2024egozero,haldar2025pointpolicy}, we anchor
each \geo{} to an entity, so that ${}^{E}\!T_{LH}$ and
${}^{E}\!T_{RH}$ track the hand--object configuration as it evolves
through approach, grasp, and transport, making the representation
\emph{interaction-centric} by construction.
Expressing them relative to scene entities rather than the camera makes
these hand--entity blocks exactly viewpoint-invariant, enabling direct
human-to-robot transfer; the entity's own pose
${}^{\mathrm{REF}}\!T_{E}$ remains tied to the reference frame, so
robustness to novel viewpoints comes from demonstrations spanning many
head poses rather than from exact invariance.
We also gain a unified, variable-length interface that accommodates
scenes with different numbers of objects without architectural
changes.
We empirically show that \geo{} is the key enabler of cross-embodiment
transfer (Sec.~\ref{sec:generalization},~\ref{sec:ablations}).

\subsection{Flow Matching Policy with Dense Auxiliary Objectives}
\label{sec:policy}

Our policy (Fig.~\ref{fig:system_overview}) takes the scene state
$s_t$---\geo{} tokens and an RGB image---and generates a bimanual
action trajectory $\mathbf{a} \in \R^{K \times D_a}$ over a $K$-step
horizon, where each $D_a$-dim slice concatenates both hands' 6-DoF
poses and binary grasps. We describe the training below.

\textbf{Flow matching action generation.}
We formulate action generation as a conditional flow
matching~\citep{lipman2023flow,liu2023flow} problem: we parameterize
a velocity field $v_\theta$ with a transformer decoder conditioned on
$s_t$, and train it to transport a Gaussian prior sample to the action
target.
Our primary training loss is:
\begin{equation}
    \mathcal{L}_{\text{FM}} = \mathbb{E}_{t,\, \mathbf{x}_0,\,
    \mathbf{x}_1}
    \Big[
    w_p\left\| \Delta\mathbf{p} \right\|^{2} +
    w_r\left\| \Delta\mathbf{r} \right\|^{2} +
    w_g\left\| \Delta g \right\|^{2}
    \Big],
    \label{eq:fm_loss}
\end{equation}
where $w_p, w_r, w_g$ are the loss weights for
position~($\mathbf{p}$), rotation~($\mathbf{r}$), and grasp~($g$);
$\Delta(\cdot) = v_\theta(\mathbf{x}_t, t, s_t) - (\mathbf{x}_1 - \mathbf{x}_0)$
is the velocity prediction error;
$\mathbf{x}_t = (1{-}t)\mathbf{x}_0 + t\mathbf{x}_1$ is the
interpolated sample at flow time $t \sim \mathcal{U}(0,1)$;
$\mathbf{x}_0 \sim \mathcal{N}(\mathbf{0}, \mathbf{I})$ is a Gaussian
prior sample; and $\mathbf{x}_1$ is the ground-truth bimanual action.
At inference, we integrate the learned ODE with a fixed-step Euler
solver.

\textbf{Dense auxiliary objectives.}
To extract rich supervision from every demonstration, we add three
auxiliary objectives that share the context encoder with the flow
matching head:
(1)~\emph{Object motion} ($\mathcal{L}_{\text{OM}}$): we predict each
manipulated object's future 6-DoF trajectory, forcing the encoder to
model object dynamics under hand motion;
(2)~\emph{2D trace} ($\mathcal{L}_{\text{2D}}$): we regress future 2D
projections of entity trajectories, grounding the representation in
the visual observation;
(3)~\emph{Latent consistency} ($\mathcal{L}_{\text{LC}}$): we
predict the \geo{} state $K$ steps ahead, pushing the encoder to
capture scene dynamics.
We combine them with the flow matching loss into a single objective:
\begin{equation}
    \mathcal{L} = \mathcal{L}_{\text{FM}}
    + \lambda_{\text{OM}}\,\mathcal{L}_{\text{OM}}
    + \lambda_{\text{2D}}\,\mathcal{L}_{\text{2D}}
    + \lambda_{\text{LC}}\,\mathcal{L}_{\text{LC}},
    \label{eq:total_loss}
\end{equation}
where $\lambda_{\text{OM}}, \lambda_{\text{2D}}, \lambda_{\text{LC}}$
are the loss weights of the three auxiliary objectives.
We derive every auxiliary target automatically from the perception
pipeline, so each demonstration yields a dense multi-task signal.
All three objectives forecast how the scene evolves in complementary
spaces (3D physical, 2D visual, latent space), equipping the shared
encoder with a lightweight world model of hand--object interaction.
We also exploit the shared encoder as a multi-task regularizer that
curbs overfitting, with the largest gains in the low-data regime
(Sec.~\ref{sec:data_efficiency},~\ref{sec:ablations}).

%-----------------------------------------------------------------------
\subsection{Deployment on a Robot}
\label{sec:deployment}
Nothing inside the policy changes when it moves from human video to a
robot: it consumes the same two inputs as in training---one
embodiment-agnostic RGB image and a set of \geo{} tokens---and only the
modules that \emph{produce} them are swapped.
At deployment the camera is static, so its frame is the reference frame
$\mathrm{REF}$ directly; on human video we instead freeze the camera pose
at the first manipulation frame, so each demonstration contributes its own
$\mathrm{REF}$ and a new camera placement is another draw from that
distribution rather than an out-of-distribution input.
The robot-side frontend is the training frontend with triangulation
replaced by depth back-projection, needing neither SLAM nor the scene
sweep: detection, segmentation, and orientation estimation run \emph{once}
per episode from one text prompt per object, while arm forward kinematics
supplies $T_{\text{ee}}$ and $g$ every cycle.
Grasp latching carries over unchanged (Sec.~\ref{sec:spatial}) with the end
effector in place of the hand, so a grasped object keeps a fixed pose
relative to the gripper.
The loop is synchronous at ${\sim}10$\,Hz on a single RTX~4090, so each
plan acts on a frame at most ${\sim}100$\,ms old.
The only per-rig setup is camera intrinsics and one hand--eye extrinsic per
arm; we use no CAD models, no manual initialization, and no depth inside
the policy. Appendix~\ref{app:deployment} details the full pipeline.

\section{Experiments}
\label{sec:experiments}

\begin{wrapfigure}{r}{0.48\textwidth}
    \vspace{-30pt}
    \centering
    \includegraphics[width=0.46\textwidth]{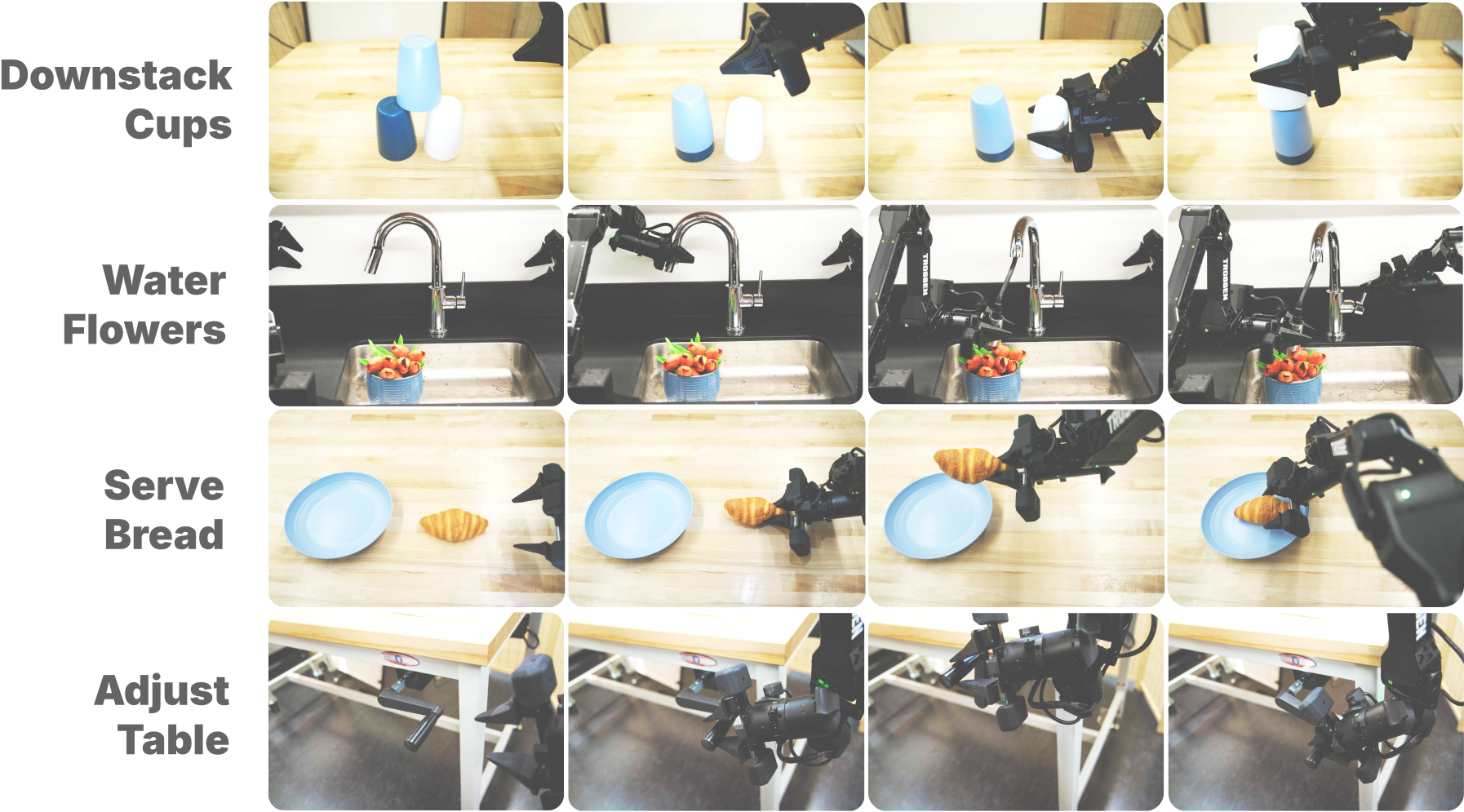}
    \caption{\textbf{Four Real-World Evaluation tasks.}}
    \label{fig:tasks}
    \vspace{4pt}
\end{wrapfigure}

We evaluate \ours{} to answer four questions:
(1)~Can the embodiment gap be bridged to achieve reliable manipulation
from human video alone? (Sec.~\ref{sec:main_results})
(2)~How does policy performance scale with human data versus matched
robot data? (Sec.~\ref{sec:data_efficiency})
(3)~How robust is the policy to distribution shifts in embodiment,
viewpoint, and environment? (Sec.~\ref{sec:generalization})
(4)~How much does each component contribute to the final
performance? (Sec.~\ref{sec:ablations})
Unless otherwise noted, all experiments are conducted on Trossen
WidowX arms with a top-mounted RealSense D405.

%-----------------------------------------------------------------------

\subsection{\ours{} Bridges the Embodiment Gap Efficiently}
\label{sec:main_results}

\begin{figure}[!t]
    \centering
    \includegraphics[width=0.90\textwidth]{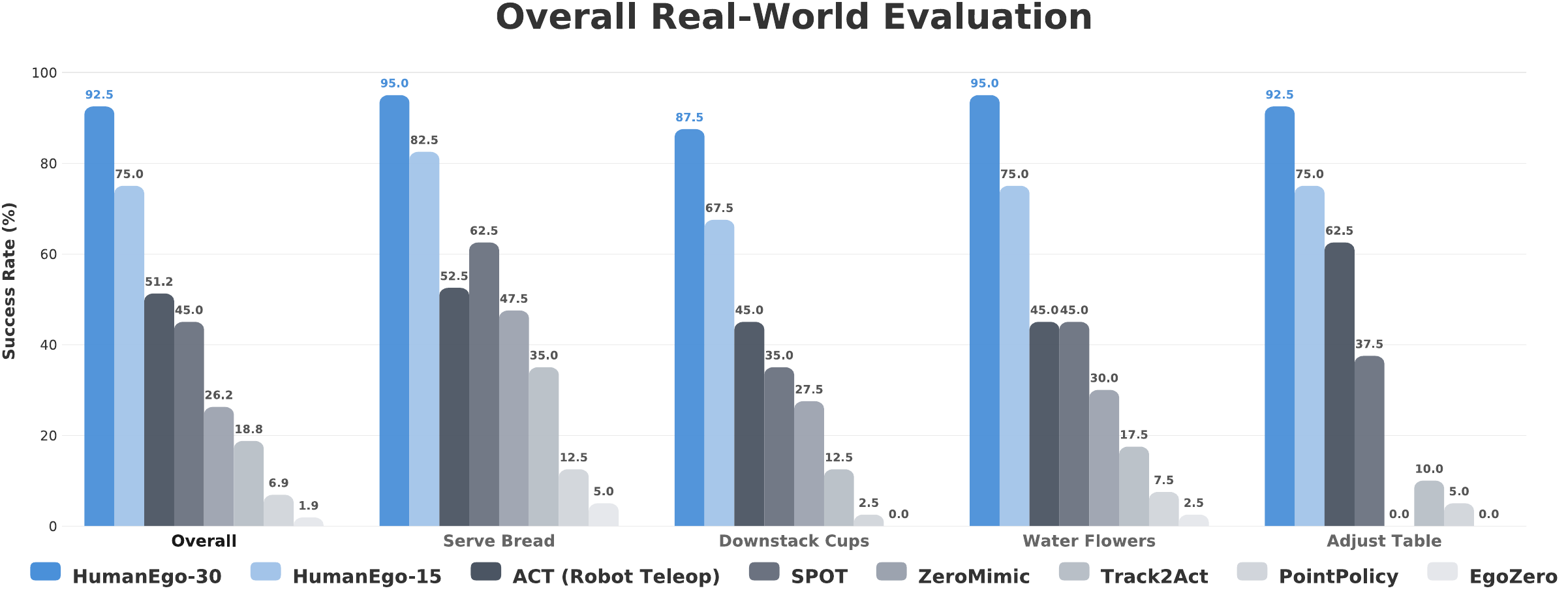}
    \caption{\textbf{Overall Real-World Evaluation.} Real-world success rate~(\%) for each method across all four tasks. \ours{} with 30~min of data achieves the highest success rate on every task, demonstrating consistent improvements over both human-video baselines and robot teleoperation methods.}
    \label{fig:baselines}
\end{figure}

We evaluate \ours{} on four real-world manipulation tasks
(Fig.~\ref{fig:tasks}):
\emph{Serve Bread}, a \textbf{pick-and-place} task in which the
robot grasps a croissant from arbitrary positions and places it on
a plate;
\emph{Downstack Cups}, a \textbf{long-horizon multi-step} task
requiring sequential toppling, grasping, and re-stacking of three
nested cups;
\emph{Water Flowers}, a \textbf{contact-rich bimanual} task with
strict temporal ordering---one arm holds a pulled-out spray head
over a flower pot while the other opens the valve; and
\emph{Adjust Table}, a \textbf{sustained rotational-control} task
in which the robot grasps a crank handle and turns it three full
revolutions without releasing.
Every trial starts from a randomized initial object position, and we
report success rate~(\%) over 40~trials per task.
We compare against five zero-shot methods that learn manipulation from
human egocentric video, spanning three design families: point-based
(\emph{EgoZero}~\citep{ze2024egozero},
\emph{Point Policy}~\citep{haldar2025pointpolicy}), goal-conditioned
(\emph{ZeroMimic}~\citep{he2024zeromimic},
\emph{Track2Act}~\citep{bharadhwaj2024track2act}), and object-centric
(\emph{SPOT}~\citep{hsu2024spot}); Appendix~\ref{app:baselines} describes
each. All five are
trained on the same 30~minutes of human data per task, are
reimplemented in our codebase, and receive the same perception
frontend, controller, and post-processing as \ours{}, including the
grasp latching of Sec.~\ref{sec:spatial}, so differences reflect the representation and policy rather
than the surrounding system. We also
include \emph{ACT}~\citep{zhao2023act}, trained on 30~minutes of
robot teleoperation collected on the same hardware.
A human demonstration takes about 40\,s, beginning with a
${\sim}1$\,s scene sweep that supplies the parallax for triangulation,
whereas an experienced operator needs about 60\,s per teleoperated
demonstration.

\textbf{\ours{} achieves the highest success rate on every single
task.}
As shown in Fig.~\ref{fig:baselines}, \ours{} with only 30~minutes of
human data reaches \textbf{92.5\%} average success rate across all four
tasks.
In contrast, the five human-video baselines range from 1.9\% to 45.0\%:
each captures only a partial aspect of manipulation, performing adequately
on simpler tasks but collapsing on those that demand precise hand--object
reasoning. The lowest scores come from the point-based methods, which
encode the scene as unstructured points in absolute coordinates with no
entity separation and no relative hand--object transform; their failures
concentrate at the grasp, where the gripper arrives offset from the
randomized object and the episode cannot recover.
\ours{} is the only method that maintains high performance regardless
of task complexity.

\textbf{Even with half the data, \ours{} outperforms robot
teleoperation.}
\ours{} with only 15~minutes of human data already reaches 75.0\%,
surpassing ACT trained on 30~minutes of robot teleoperation (51.2\%),
highlighting the data efficiency of our approach---and that
\emph{readily available human videos can be a surprisingly potent data
source for robot learning.}

\textbf{\ours{} excels on tasks that demand precise coordination and
spatial reasoning.}
On Downstack~Cups, where ${\sim}1$\,cm tolerance makes early errors
compound, \ours{} reaches 87.5\% while no baseline exceeds 45\%; on
Water~Flowers, which demands strict bimanual ordering and precise aiming
of the stream into the pot, it achieves 95\%, more than double the best
baseline (45\%).

%===============================================================================
% Baselines figure — Page 6 (full width)
%===============================================================================

%-----------------------------------------------------------------------

\subsection{How Does Performance Scale with Human Data?}
\label{sec:data_efficiency}

\begin{wrapfigure}{r}{0.48\textwidth}
    \vspace{-6pt}
    \centering
    \includegraphics[width=0.46\textwidth]{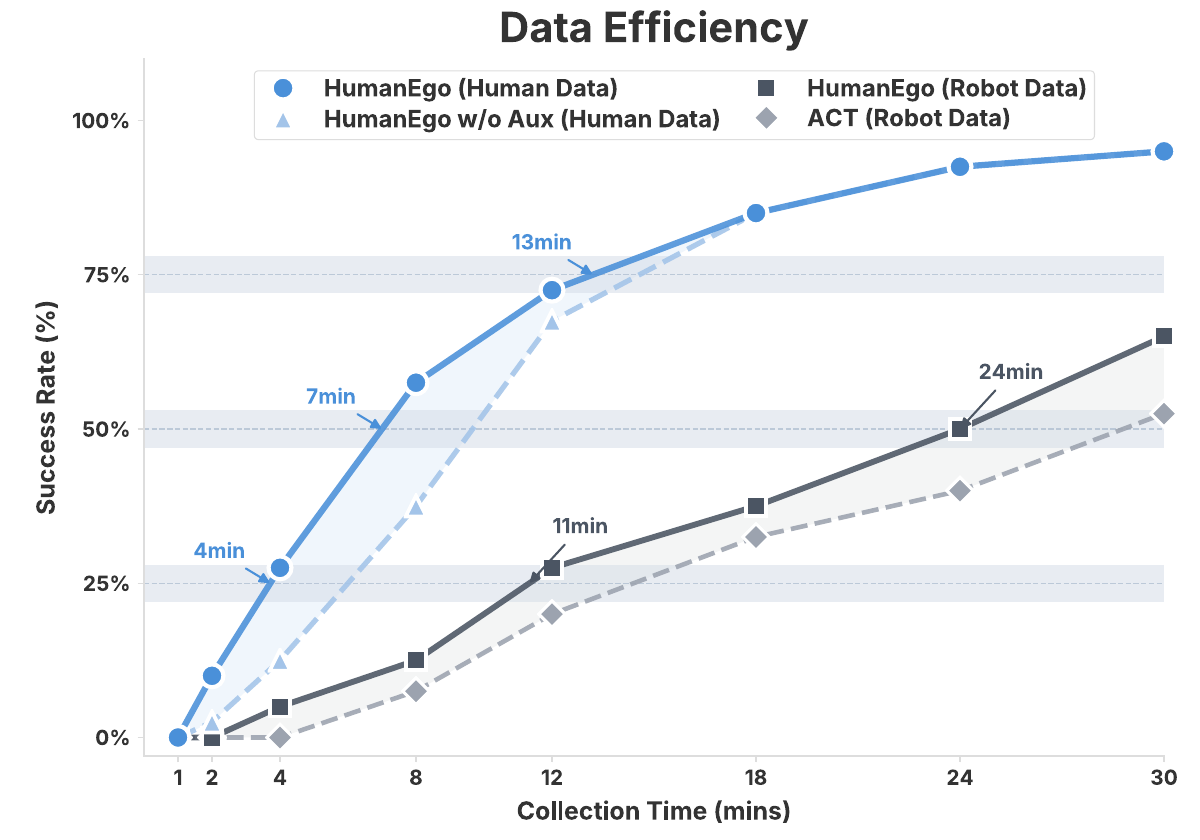}
    \caption{\textbf{Data efficiency.} Success rate~(\%) vs.\ data collection time. \ours{} trained on 8~min of human data surpasses ACT's 30-min robot data.}
    \label{fig:scaling}
    \vspace{-10pt}
\end{wrapfigure}

We compare \ours{} trained on human video against ACT and \ours{}
trained on robot teleoperation as a function of collection time on
Serve Bread (Fig.~\ref{fig:scaling}).

\textbf{\ours{} learns effectively from minimal human data.}
With only $\sim$7~minutes of human demonstrations, \ours{} already
reaches 50\% success rate and continues to climb smoothly, reaching
95\% at 30~minutes.
This steep, monotonic scaling curve indicates that our pipeline
extracts useful manipulation signal from even a handful of
demonstrations.

\textbf{Auxiliary objectives amplify learning when demonstrations are
scarce.}
Between 2 and 12~minutes, \ours{} with auxiliary losses consistently
outperforms the variant without them, with the largest gain at 8~min
(57.5\% vs.\ 37.5\%).
At 18~minutes and beyond the two variants coincide, both reaching 95\% at
30~minutes, confirming that
auxiliary losses extract richer supervision from each demonstration---a
benefit that diminishes as data grows abundant.

\begin{wrapfigure}{r}{0.48\textwidth}
    \vspace{-10pt}
    \centering
    \includegraphics[width=0.46\textwidth]{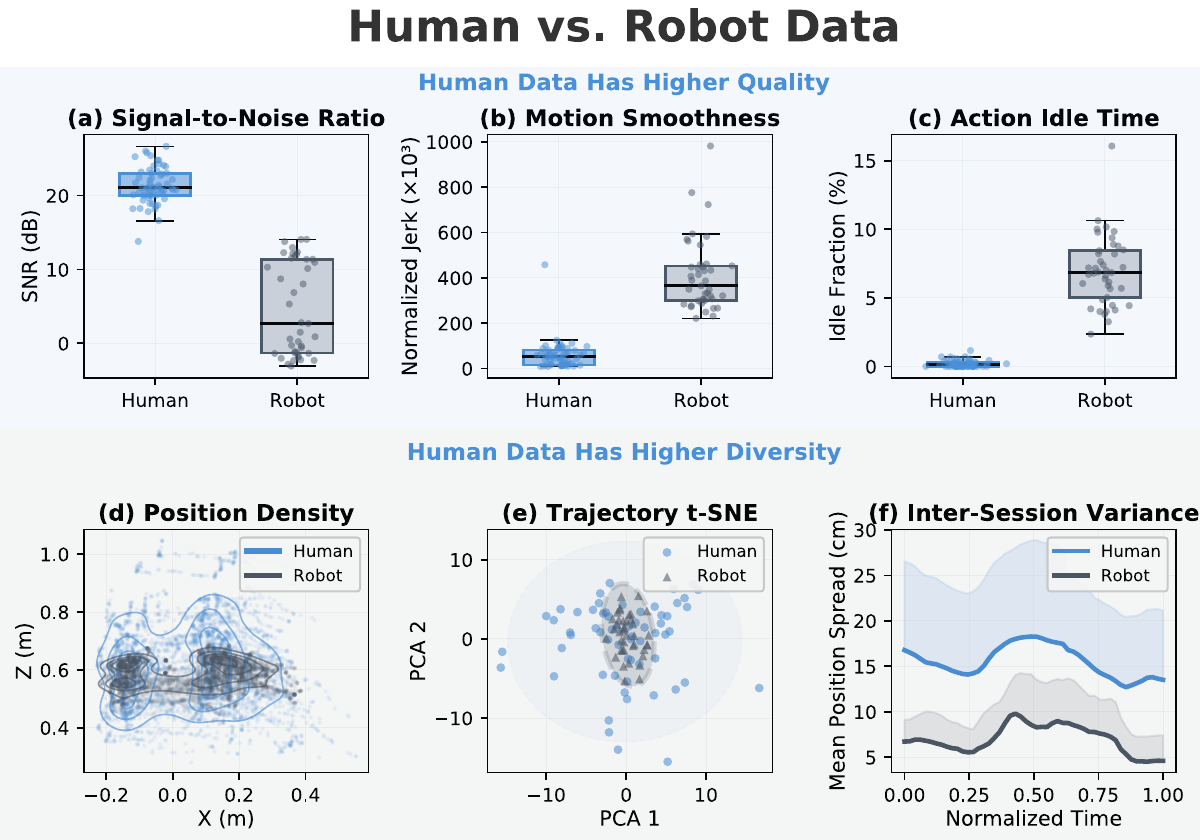}
    \caption{\textbf{Human vs.\ robot data.} Human egocentric data exhibits higher SNR, smoother motion, less idle time (top), and greater spatial and trajectory diversity (bottom).}
    \label{fig:data_comparison}
    \vspace{-10pt}
\end{wrapfigure}

\textbf{Human video is a more efficient data source than
robot teleoperation.}
At 8~minutes of collection time, \ours{} trained on human video
(57.5\%) already surpasses ACT trained on 30~minutes of robot
teleoperation (52.5\% on this task)---a \textbf{3.75$\times$} reduction in collection
effort.
Holding the architecture fixed isolates the data source: the same policy
trained on 30~min of robot teleoperation reaches 65\% versus 95\% on human
video (\textbf{+30\,pp}), so the gain is not an artifact of the policy
class.

\textbf{Diversity, not quantity alone, drives the gap.}
Offline metrics (Fig.~\ref{fig:data_comparison}) show where the extra
signal comes from: per minute of collection, human demonstrations have
higher signal-to-noise ratio, smoother motion, and less idle time, and
they spread more widely over object positions, trajectory shapes, and
sessions. Matched minutes of teleoperation therefore buy fewer and less
varied interactions.

%-----------------------------------------------------------------------

\begin{figure}[H]
    \centering
    \begin{minipage}[t]{0.48\textwidth}
        \centering
        \includegraphics[width=0.93\textwidth]{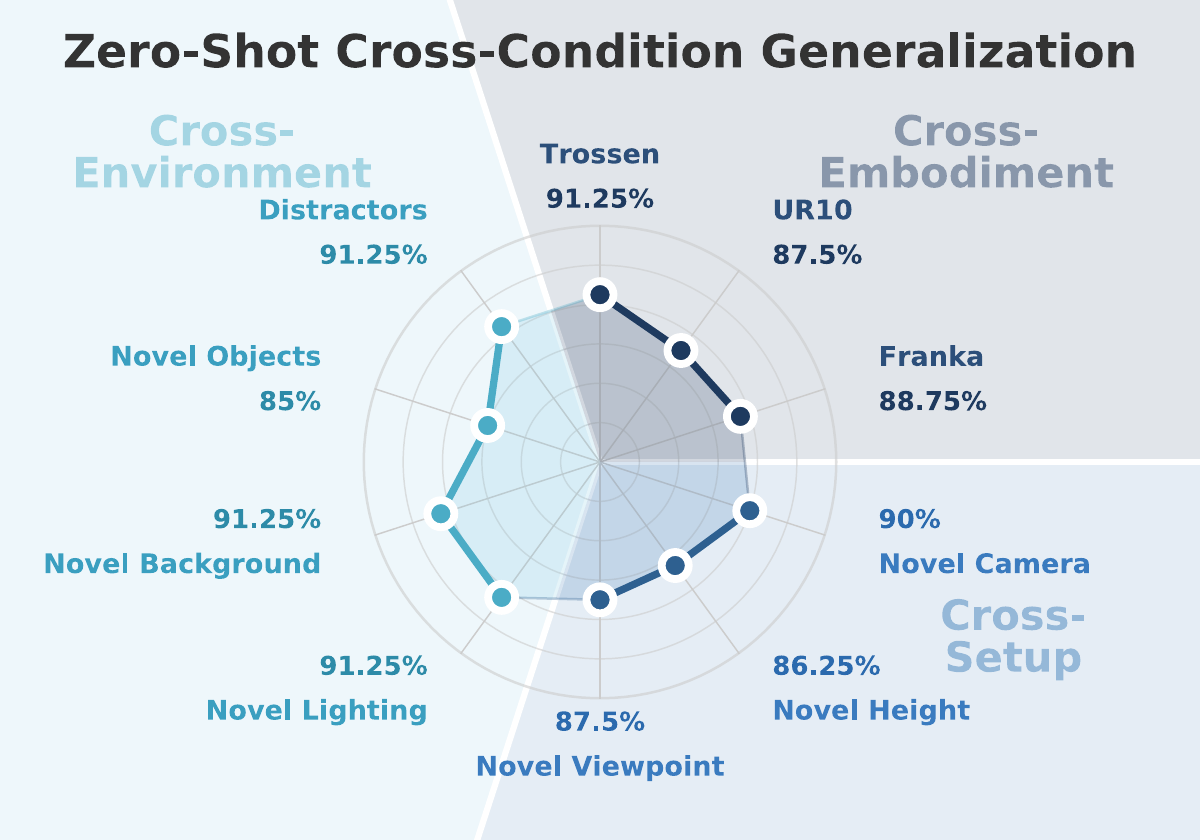}
        \caption{\textbf{Zero-Shot Cross-Condition Generalization.} \ours{} maintains robust success across different conditions without retraining.}
        \label{fig:generalization}
    \end{minipage}%
    \hfill
    \begin{minipage}[t]{0.48\textwidth}
        \centering
        \includegraphics[width=0.93\textwidth]{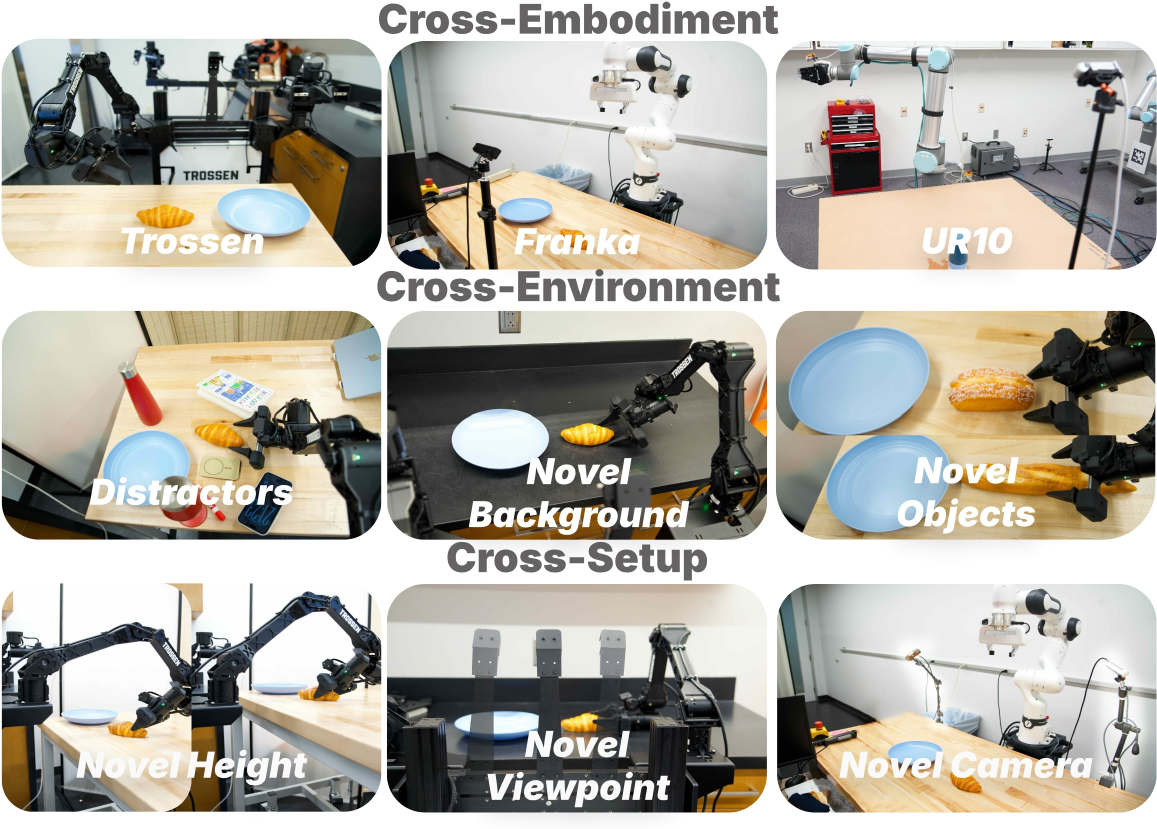}
        \caption{\textbf{Cross-condition real-world evaluation:} cross-embodiment / environment / setup.}
        \label{fig:realworld}
    \end{minipage}
\end{figure}

\subsection{One Policy, Many Conditions}
\label{sec:generalization}

We deploy \ours{} on Serve Bread and Downstack Cups, each trained on
30~minutes of human data, across the 9 out-of-distribution conditions of
Fig.~\ref{fig:generalization}---Trossen is the in-distribution
reference---with no retraining, and report the mean success rate over the
two tasks (setup in Fig.~\ref{fig:realworld}).

%===============================================================================
% Generalization + Cross-Condition — placed inline below deploy paragraph
%===============================================================================

\textbf{\ours{} is robust across the tested visual conditions.}
Changing the background, lighting, or viewpoint, adding distractors,
or swapping in object instances that never appear in the training set
all yield 85--91.25\% success, with no measurable degradation in most
cases.
Each condition varies one factor at a time, so this demonstrates that the
policy extracts task-relevant information from the visual input while
remaining insensitive to the nuisance factors we tested.

\textbf{\ours{} is robust to varied object placements.}
On Serve Bread, it delivers the bread to the plate across a wide range of
absolute and relative positions on the table and even at novel heights; on Downstack Cups, it completes the three-step sequence
under varied table heights and cup positions---scenarios where
methods like EgoZero~\citep{ze2024egozero} and
\emph{Point Policy}~\citep{haldar2025pointpolicy} often fail. This
robustness reflects the object-centric structure of our
interaction-centric representation.

\textbf{\ours{} is hardware-agnostic.}
All training data is collected with Aria glasses, yet the same
checkpoint drives a Trossen, Franka, or UR10 arm under a different
camera, sharing no hardware with the training setup.
It transfers once each arm's hand--eye extrinsic is known
(Sec.~\ref{sec:deployment})---enabled by the interaction-centric
representation detailed in Sec.~\ref{sec:spatial}.

%-----------------------------------------------------------------------

\subsection{What Drives Performance of \ours{}?}
\label{sec:ablations}

\begin{figure}[!t]
    \centering
    \begin{minipage}[t]{0.48\textwidth}
        \centering
        \includegraphics[width=0.93\textwidth]{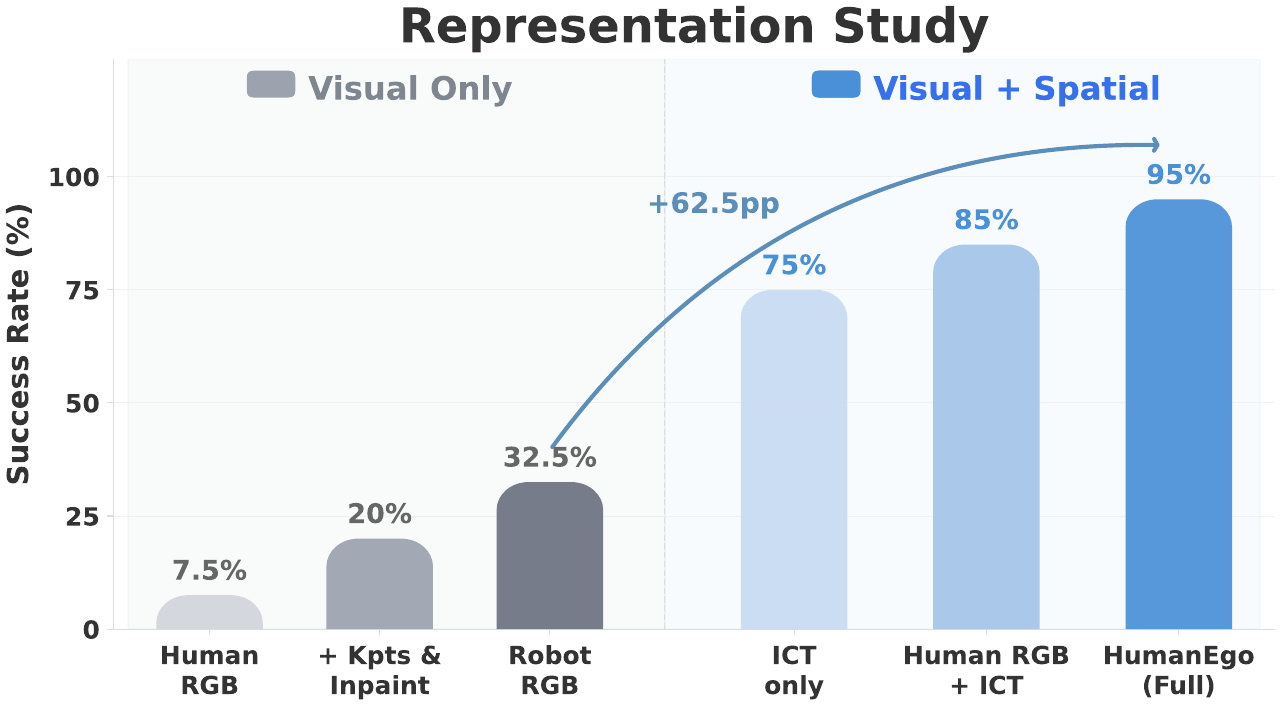}
        \caption{\textbf{Representation study.} Success rate~(\%) for six input configurations. Visual-only methods plateau at 32.5\% with any strategy, whereas the full system reaches 95\% (+62.5\,pp).}
        \label{fig:representation}
    \end{minipage}%
    \hfill
    \begin{minipage}[t]{0.48\textwidth}
        \centering
        \includegraphics[width=0.93\textwidth]{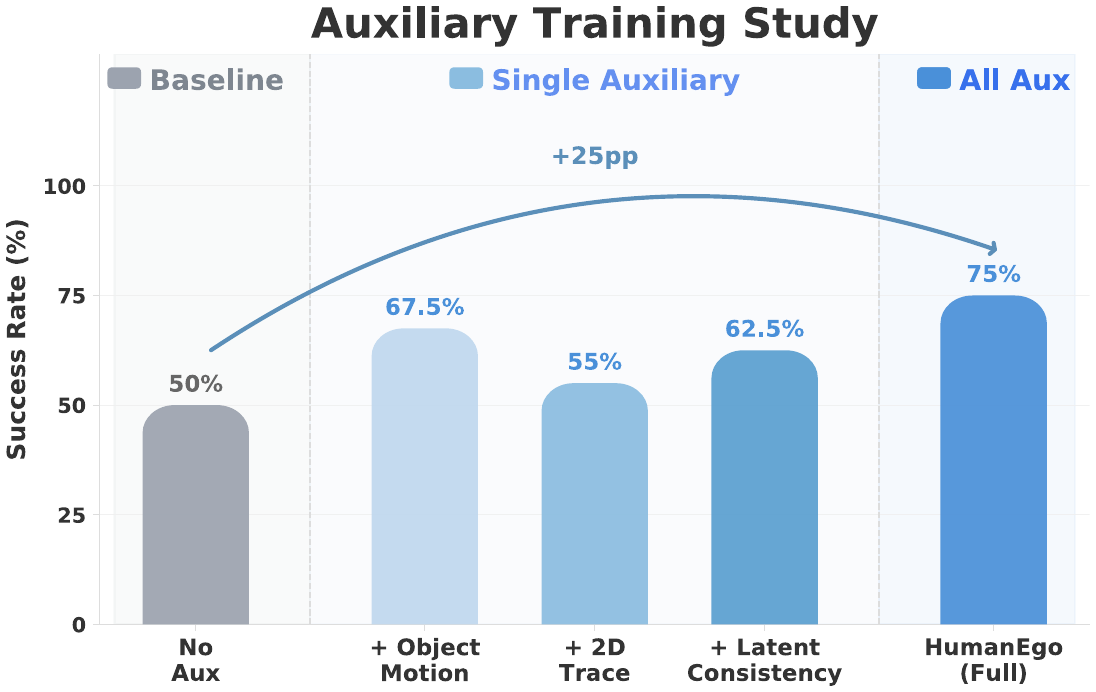}
        \caption{\textbf{Auxiliary training study.} Success rate~(\%) at 15\,min of data for each auxiliary objective individually. Each improves over the baseline; all three combined give +25\,pp.}
        \label{fig:aux_training}
    \end{minipage}
\end{figure}

We ablate \ours{}'s two core design choices on Water Flowers: the
spatial representation at 30~minutes of data
(Fig.~\ref{fig:representation}) and the auxiliary training objectives
at 15~minutes (Fig.~\ref{fig:aux_training}).

\textbf{Explicit spatial representation, not visual fidelity alone, is
the key to bridging the embodiment gap.}
We isolate visual preprocessing from spatial representation.
Progressively reducing the visual embodiment gap---from raw human RGB
(7.5\%) to keypoint rendering with arm inpainting (20\%) to robot RGB
that eliminates the gap entirely (32.5\%)---yields only modest gains;
even with zero visual mismatch, the policy barely exceeds 30\%.
Monocular RGB encodes \emph{appearance}, not the 3D spatial
relationships that manipulation demands.
Adding \geo{} to raw human RGB produces a dramatic jump from 7.5\% to
85\%, and the full system reaches 95\%.
In the other direction, removing the image branch so that visual
information reaches the policy only through \geo{} yields 75\%: the
image stream contributes 20\,pp on top of \geo{}. Since \geo{} is itself derived from vision, this variant routes
visual information through entities rather than removing it.
\geo{} directly encodes the relative 6-DoF transforms between hands and
objects---the core manipulation state---turning the problem of inferring
3D dynamics from pixels into learning actions from explicit spatial
relationships.

\textbf{Auxiliary objectives provide complementary gains.}
We evaluate each objective individually at 15~minutes of data.
Each one improves over the no-auxiliary baseline---object motion
($+$17.5\,pp), 2D trace ($+$5\,pp), and latent consistency
($+$12.5\,pp)---and combining all three yields $+$25\,pp.
Together, the ablations support the main experimental thesis:
\geo{} supplies the transferable manipulation state, and dense
auxiliary supervision makes that state learnable from minutes of
human video.

%===============================================================================
% Representation + Aux Training — Page 8
%===============================================================================

%===============================================================================

\section{Conclusion}
\label{sec:conclusion}

We presented \textbf{\ours{}}, a framework that learns robot manipulation
policies from minutes of human egocentric video---without any robot data or
large-scale pretraining. A hardware-agnostic representation bridges the
embodiment gap through visual preprocessing (arm inpainting, keypoint
rendering) and spatial encoding (\geo{}), while a flow matching policy with
dense auxiliary objectives learns from scarce data. \ours{} reaches 92.5\%
average success across four real-world tasks, improves on matched-time robot
teleoperation by 41 percentage points, and transfers zero-shot to novel
embodiments, cameras, and environments without retraining.

\textbf{Limitations and future work.}
\ours{} relies on Aria's stereo hand tracking; monocular substitutes drop
real-world success sharply. Object poses are estimated once per episode and
then propagated kinematically through the hand---or the gripper at
deployment---so we assume a quasi-static scene: in-hand reorientation, slip,
or re-grasping violates it. Retargeting maps the thumb--index pair onto a
parallel-jaw pose and a binary grasp, so transfer is demonstrated across
parallel-jaw arms rather than dexterous hands.
Off-the-shelf perception failures also cascade into the policy, and
few-shot learning plateaus at ${\sim}$1\,cm precision. Appendix~\ref{app:limitations} expands these and outlines future
work.

%===============================================================================

%===============================================================================

%==============================================================================

\bibliography{example}  % .bib

%===============================================================================
% Appendix (does not count toward page limit for CoRL)
%===============================================================================
\clearpage
\appendix
\section*{Appendix}

%-----------------------------------------------------------------------
\section{Extended Related Work}
\label{app:related}

Recent years have produced rich large-scale egocentric and
hand--object interaction
datasets~\citep{grauman2022ego4d,damen2022epickitchens100,hoque2026egodex,banerjee2024hot3d,liu2022hoi4d,liu2024taco,wang2023holoassist,punamiya2026egoverse}
that provide the data foundation for learning manipulation from
human video. Building on this foundation, one line of work scales
up generalist policies and world
models~\citep{yang2025egovla,zheng2026egoscale,zhang2026unidex,lee2025tracegen,yuan2024general}
that learn embodiment-agnostic representations from massive
corpora, yet deployment demands enormous compute and per-task robot
post-training. Another line
co-trains~\citep{kareer2024egomimic,punamiya2025egobridge,jain2024vid2robot,zhu2025emma,liu2025immimic,kareer2025human2robot,qiu2025humanoid}
on paired human and robot data, jointly optimizing across
embodiments to amortize the human signal, yet every new task still
requires its own batch of robot demonstrations. Visual retargeting
approaches~\citep{lepert2025phantom,wang2024h2r,lepert2025masquerade,dessalene2025embodiswap}
synthesize pseudo-robot demonstrations by inpainting the human arm
and rendering a robot in its place, but the rendered imagery is
brittle to morphological and viewpoint variations. Hierarchical
methods~\citep{wang2023mimicplay,xu2023xskill,kim2025uniskill} learn high-level plans from human
video and delegate low-level control to a robot-trained controller,
which still depends on robot data for the low-level skill. To avoid
robot data altogether, a third family pursues embodiment-agnostic
representations for zero-shot transfer, differing in \emph{what}
they represent: point-based
methods~\citep{ze2024egozero,haldar2025pointpolicy,guzey2025aina,singh2025afford2act}
encode the scene as 2D or 3D points and gain computational
efficiency, but lose the structural relationship between hand and
object; object-centric
methods~\citep{hsu2024spot,zou2026activeglasses,patel2025rigvid,yu2025genflowrl,yin2025motionfield}
represent the scene through the object's 6-DoF pose or motion,
capturing object dynamics yet modeling the manipulator only
implicitly; and goal-conditioned
methods~\citep{bharadhwaj2024track2act,he2024zeromimic} predict 2D
tracks or 3D wrist trajectories conditioned on a target image, but
require explicit goal specification at deployment. Several other
directions~\citep{park2025demodiffusion,xu2024im2flow2act,shah2025mimicdroid,li2025novaflow,chen2026videomanip,shi2025points2reward,planner,watch}
also explore learning manipulation from video along complementary
axes. A common thread runs
through these zero-shot lines: they represent the hand \emph{or}
the object, but rarely their \emph{interaction}---the very signal
that defines manipulation. \ours{} bridges this gap with an
interaction-centric representation that explicitly encodes the
spatial relationship between hands and objects, achieving zero-shot
transfer from only minutes of human egocentric video without any
robot data or large-scale pretraining.

%-----------------------------------------------------------------------

%-----------------------------------------------------------------------
\section{Extended Limitations and Future Work}
\label{app:limitations}

\textbf{Limitations and future work.}
Our framework relies on Aria's stereo hand tracking---monocular
substitutes drop real-world success sharply, calling for stronger monocular
hand pose estimators that recover absolute depth.
Object poses are estimated once at the start of an episode and
thereafter tracked kinematically through the gripper, so we assume a
quasi-static scene: unattended objects do not move on their own, and a
grasped object does not shift in the hand. In-hand reorientation, slip,
or re-grasping violates this assumption and would require online,
occlusion-robust tracking.
The pipeline chains several off-the-shelf perception modules whose
failures cascade into the policy, motivating stronger or
jointly-trained frontends.
Finally, few-shot learning plateaus at ${\sim}$1\,cm precision;
reaching sub-centimeter accuracy on contact-rich tasks will
likely require reinforcement-learning refinement or
simulation-based fine-tuning.
We see \ours{} as a starting point---a fully zero-shot,
robot-data-free pipeline that future work can extend with online
tracking, learned frontends, and downstream RL refinement.

%-----------------------------------------------------------------------
\section{Baseline Implementations}
\label{app:baselines}

The five zero-shot human-video baselines in Sec.~\ref{sec:main_results}
span three design families.
Two are point-based: \emph{EgoZero}~\citep{ze2024egozero} lifts the
egocentric scene into 3D point clouds that treat hand and object alike,
while \emph{Point Policy}~\citep{haldar2025pointpolicy} instead tracks
sparse keypoints on each.
Another two are goal-conditioned:
\emph{ZeroMimic}~\citep{he2024zeromimic} distills 3D wrist trajectories
from web video and conditions on a goal image at test time, and
\emph{Track2Act}~\citep{bharadhwaj2024track2act} predicts 2D point tracks
without explicit 3D reasoning.
The last, \emph{SPOT}~\citep{hsu2024spot}, is object-centric, generating
\SE{} object trajectories with a diffusion model.

\section{Data Collection Details}
\label{app:data_collection_details}

\subsection{Aria Gen1 Glasses}
\label{app:aria_setup}

\begin{wrapfigure}{r}{0.33\textwidth}
    \vspace{-14pt}
    \centering
    \includegraphics[width=0.31\textwidth]{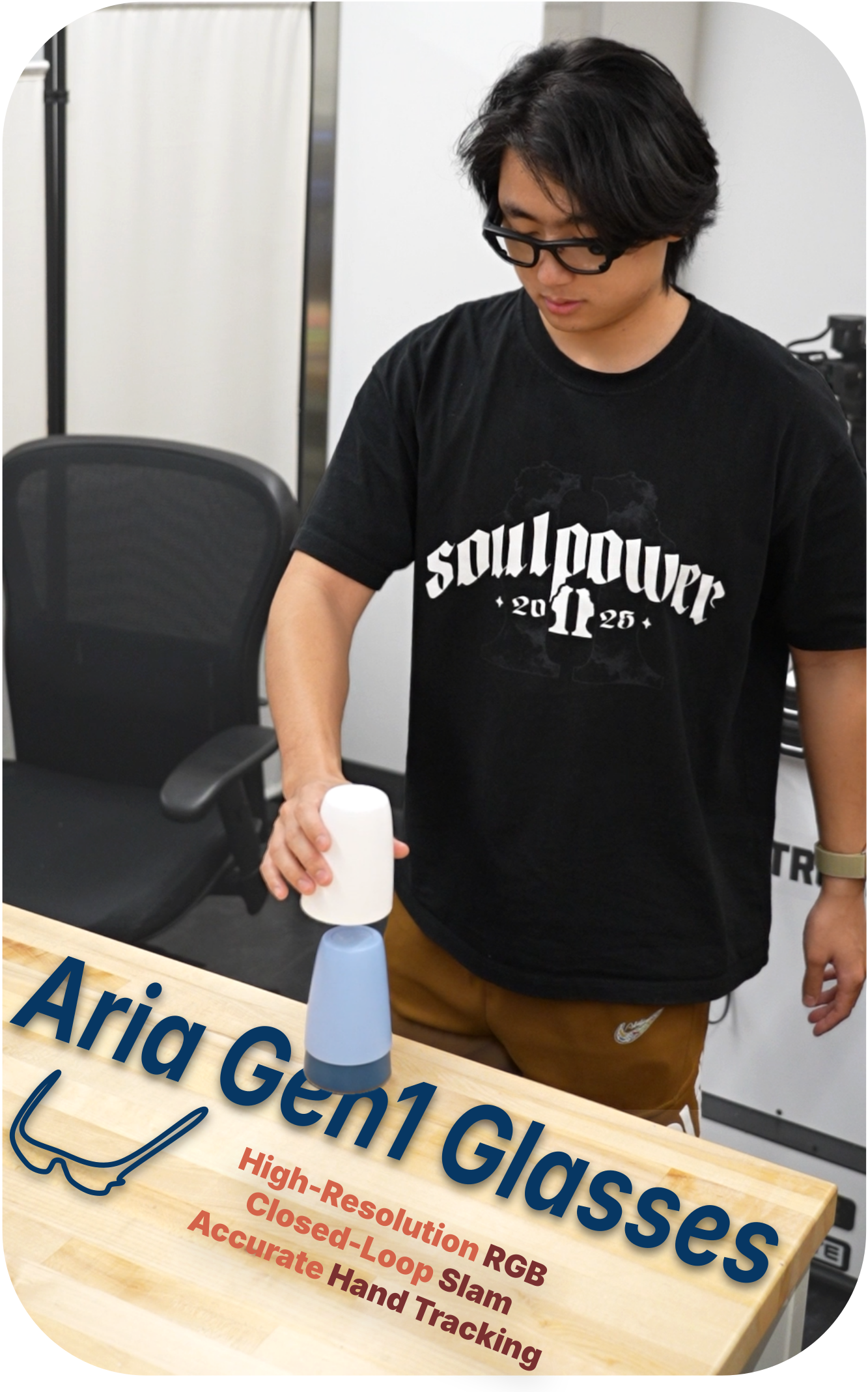}
    \caption{\textbf{Data collection setup.}}
    \label{fig:datacollection}
    \vspace{-10pt}
\end{wrapfigure}

\paragraph{Aria Gen1 recording configuration.}
We record every human demonstration with Project Aria Gen1 glasses
(Fig.~\ref{fig:datacollection}),
configured through the official Project Aria Mobile App with the
sensor profile listed below:
\begin{itemize}[itemsep=1pt, topsep=2pt, leftmargin=15pt, parsep=0pt]
  \item \textbf{RGB:} $30$\,fps at $2$\,MP.
  \item \textbf{SLAM:} $2{\times}$ monochrome cameras, $30$\,fps at
    VGA.
  \item \textbf{ET (eye tracking):} $2{\times}$ cameras, $10$\,fps
    at QVGA.
  \item \textbf{IMUs:} two $6$-axis IMUs sampled at $1000$\,Hz and
    $800$\,Hz.
  \item \textbf{Magnetometer, barometer, $7$-mic audio array, GPS,
    Wi-Fi, and BLE:} all enabled for synchronization metadata and
    environmental context.
\end{itemize}
All streams are hardware-timestamped on-device and synchronized to a
common Aria clock, so every modality is time-aligned at the
millisecond level.

\paragraph{Aria Machine Perception Services.}
On top of the raw recordings, Aria's cloud-hosted \emph{Machine
Perception Services} (MPS)~\citep{engel2023aria} post-process each
capture into metric, ready-to-use signals. Two MPS outputs are
critical for our pipeline:
\begin{itemize}[itemsep=1pt, topsep=2pt, leftmargin=15pt, parsep=0pt]
  \item \emph{Closed-loop trajectory.} The closed-loop SLAM output
    fuses the two monochrome SLAM cameras and the two IMUs and
    applies loop closure plus global optimization to yield a
    globally consistent, drift-corrected 6-DoF device trajectory in
    a gravity-aligned world frame. We query this trajectory at every
    RGB frame to obtain the calibrated camera extrinsics used for
    triangulation (App.~\ref{app:triangulation}) and to lift every
    hand keypoint into the world frame consumed by \geo{}
    (Sec.~\ref{sec:spatial}).
  \item \emph{Hand tracking.} The MPS hand tracker jointly processes
    the stereo SLAM cameras to produce $21$ 3D keypoints per hand
    (five per finger plus the wrist), reported directly in the world
    frame with per-keypoint confidence scores. This supplies the 3D
    hand skeleton consumed by our hand-to-gripper retargeting
    (App.~\ref{app:hand2gripper}).
\end{itemize}
Together, the closed-loop trajectory and the hand tracker provide a
metric 6-DoF camera pose and a 3D hand skeleton at every frame,
without any calibration or scene instrumentation beyond wearing the
glasses.

\subsection{Task Details}
\label{app:task_details}

We evaluate \ours{} on four real-world manipulation tasks spanning
pick-and-place, multi-step bimanual coordination, contact-rich
reasoning, and sustained rotational control (Fig.~\ref{fig:tasks}).
For each task we describe the scene, per-trial randomization, target
behavior, and success/failure criteria used to score the 40 trials
per condition reported in the main paper.

\paragraph{Serve Bread.}
\emph{Scene.} A croissant and a dinner plate sit on a tabletop, plate
on the left and bread on the right.
\emph{Randomization.} Across trials we independently randomize
(i)~the horizontal offset between the two objects, sampled from
$[0, 50]$\,cm, and (ii)~their depths along the table's front--back
axis, so the pair is generally \emph{not} colinear.
The right arm starts from an arbitrary, non-aligned pose above the
bread.
\emph{Target behavior.} Grasp the croissant from above, lift it clear
of the table, transport it to a release pose above the center of the
plate, and open the gripper.
\emph{Success.} The croissant comes to rest on the plate, in any
orientation.
\emph{Failure.} (a)~the bread ends up outside the plate (on the
table, draped over the rim, or knocked off); or (b)~the bread slips
from the gripper during transport.

\paragraph{Downstack Cups.}
\emph{Scene.} Three cups of distinct colors on a tabletop: a
\emph{white} cup at bottom-right, a \emph{dark-blue} cup at
bottom-left, and a \emph{light-blue} cup stacked on top of the white
cup. The table height and the absolute position of the cup group are
varied across trials.
\emph{Randomization.} The horizontal gap between the white and
dark-blue cups is drawn from $[0, 1]$\,cm; the light-blue cup is
jittered left/right within $[0, 2]$\,cm on top of the white cup.
\emph{Target behavior.} A three-step sequence:
(1)~\emph{Topple}---knock the light-blue cup sideways so it lands on
top of the dark-blue cup;
(2)~\emph{Grasp}---swing over to the white cup and grasp it from
above;
(3)~\emph{Cover}---lower the white cup onto the dark-blue/light-blue
stack and release, forming a three-cup tower on the left (dark-blue /
light-blue / white, bottom-to-top).
\emph{Success.} The three cups end in the intended stable stack on
the left.
\emph{Failure.} (a)~the light-blue cup is never contacted;
(b)~it is toppled but does not land on the dark-blue cup;
(c)~the white cup is not grasped; or
(d)~the white cup is not correctly placed on the stack (misses it,
topples it, or lands off-axis so the tower collapses).
Early errors compound, so the policy must succeed at every
sub-stage.

\paragraph{Water Flowers.}
\emph{Scene.} A wall-mounted faucet stands at the front of the
workspace, with a sunken sink recessed ${\sim}10$\,cm below the
tabletop directly underneath. A flower pot filled with fresh flowers
sits inside the sink.
\emph{Randomization.} The pot is placed at one of three qualitatively
different positions in the sink (top-left, middle, or bottom-right);
the faucet is free to rotate about its vertical axis within
$[-15^{\circ}, +15^{\circ}]$.
\emph{Target behavior.} Coordinated bimanual execution with strict
temporal ordering.
\textbf{Left arm (spray head):} grasp the pull-out spray head, pull
it ${\sim}15$\,cm downward out of the faucet socket, and hold the
nozzle $3$--$5$\,cm above the center of the flower pot, pointed
downward.
\textbf{Right arm (handle):} remain visible in a stationary
pre-grasp pose near the faucet handle while the left arm works;
once the left arm is in place, grasp the handle and flick it to the
right by $3$--$5$\,cm to open the valve, so water flows onto the
flowers.
\emph{Success.} The left arm holds the spray head over the pot while
the right arm has opened the valve and water pours onto the flowers.
\emph{Failure.}
\textit{Left arm:} (a)~the spray head is not grasped or not pulled
down; (b)~it slips during pull-out or transport; or (c)~it is not
positioned directly above the pot.
\textit{Right arm:} (a)~the handle is not grasped; or (b)~it is not
flicked far enough to open the valve, or slips before water flows.

\paragraph{Adjust Table.}
\emph{Scene.} The operator faces an adjustable table whose height is
controlled by a hand crank protruding from its side, oriented roughly
horizontally.
\emph{Randomization.} The initial angle of the crank handle about
its rotation axis is jittered by $\pm 10^{\circ}$ around horizontal.
\emph{Target behavior.} The right arm (1)~approaches the crank
handle from an arbitrary initial pose and grasps it firmly, then
(2)~performs a continuous \emph{counter-clockwise} rotation about
the crank axis, completing three full revolutions
($3 \times 360^{\circ}$) without releasing the handle.
\emph{Success.} All three revolutions are completed while the grasp
is maintained throughout.
\emph{Failure.} (a)~the handle is never grasped; or (b)~the handle
slips from the gripper during the rotation, before three full
revolutions are completed.

\paragraph{Per-demonstration collection time.}
Across all four tasks, a single human egocentric demonstration
takes approximately \textbf{30--40~seconds}, while a matched
teleoperated demonstration on the same hardware takes
\textbf{60--70~seconds}. The roughly $2\times$ gap reflects that
everyday human manipulation is naturally faster and more dexterous
than teleoperation through a piloting interface; combined with the
embodiment-agnostic representation, this makes human video a
substantially more sample-efficient training source per minute of
collection.

\paragraph{Additional tasks (demonstration-only).}
Beyond the four tasks evaluated quantitatively above, we also
collected data and trained \ours{} policies on five additional
tasks, illustrating the speed with which our pipeline can be
brought to bear on new behaviors. These tasks appear in our
supplementary demonstration videos but are \emph{not} part of the
quantitative evaluation in Sec.~\ref{sec:main_results}.

\emph{Charge Devices.} The operator places a smartwatch, a pair of
earphones, and a phone from the tabletop onto their corresponding
magnetic chargers. Each item is treated as a separate sub-task,
following a \textsc{grasp~$\to$~transport~$\to$~release-on-pad}
template; we collect and train a dedicated model for each sub-task.

\emph{Unscrew Cap.} A bimanual task: one hand grasps and stabilizes
the bottle body while the other hand grasps the cap, rotates it
through 2--3 full revolutions to disengage the thread, and lifts
the freed cap aside.

\emph{Open Door.} The operator opens a self-closing door fitted
with a lever-style handle. The acting hand grasps the lever,
rotates it clockwise to release the latch, and then pulls the door
outward against its return spring.

\emph{Open Cabinet.} A bimanual task: both hands simultaneously
grasp a pair of vertically oriented handles on a cabinet door and
pull it open along an outward arc.

\emph{Grab Tissue.} The operator pinches a single sheet of tissue
from a tissue box, lifts it upward and then translates it laterally
to fully extract it, and places it on the table.

%-----------------------------------------------------------------------
\section{Preprocessing Details}
\label{app:preprocessing_details}

\subsection{Triangulation}
\label{app:triangulation}

Aria Gen1 glasses lack a depth sensor, so we recover each object's
3D position by triangulating tracked 2D keypoints across frames,
treating the moving head-mounted camera as a multi-view system with
calibrated extrinsics given by the 6-DoF Aria MPS SLAM
pose~\citep{engel2023aria}.
This requires the object to remain stationary during the observation
window; once manipulation begins, the object is free to move.

\paragraph{Pre-episode scene sweep.}
Multi-view triangulation requires the same 3D point to be seen from
sufficiently different viewpoints, yet during manipulation the
head-mounted camera is often nearly stationary while only the hands
move, collapsing the effective camera baseline.
We therefore prefix every demonstration with a short \emph{scene
sweep}: the demonstrator keeps the scene static and slowly moves
their head for ${\sim}1$--$2$ seconds (${\sim}30$--$60$ frames),
using either a horizontal left-to-right pan or a forward walk-in
toward the object, before proceeding to the actual manipulation.

\paragraph{Multi-view triangulation from 2D tracks.}
For each object we detect it in the first sweep frame with
Grounding DINO~\citep{liu2024groundingdino}, segment it with
SAM2~\citep{ravi2024sam2}, sample $N$ keypoints on the resulting
mask, and track them through the $F$ sweep frames with
CoTracker3~\citep{karaev2024cotracker}.
Let $K \in \R^{3\times 3}$ be the RGB intrinsics, $T_i =
[R_i\,|\,t_i]$ the camera-to-world SLAM pose of frame $i$, and
$P_i = K\,[R_i^{\top}\,|\,-R_i^{\top} t_i] \in \R^{3\times 4}$ the
corresponding world-to-image projection.
A track $\{\mathbf{u}_n^{(i)}\}_{i=1}^{F}$ and the unknown 3D point
$\mathbf{X}_n \in \R^4$ (homogeneous) satisfy
$\mathbf{u}_n^{(i)} \times P_i\mathbf{X}_n = 0$, which yields two
linear equations per frame:
\begin{equation}
  \begin{bmatrix}
    u_n^{(i)}\,\mathbf{p}_3^{(i)\top} - \mathbf{p}_1^{(i)\top} \\[2pt]
    v_n^{(i)}\,\mathbf{p}_3^{(i)\top} - \mathbf{p}_2^{(i)\top}
  \end{bmatrix}
  \mathbf{X}_n \;=\; \mathbf{0},
  \label{eq:triangulation}
\end{equation}
where $\mathbf{p}_j^{(i)\top}$ is the $j$-th row of $P_i$.
Stacking \eqref{eq:triangulation} across all $F$ frames gives a
$2F \times 4$ system $A_n \mathbf{X}_n = \mathbf{0}$; we solve it in
the least-squares sense via SVD by taking the right singular vector
of $A_n$ with the smallest singular value and dehomogenizing to
recover $\mathbf{x}_n \in \R^3$.
The object position is then the centroid
$\mathbf{p}_{\text{obj}} = \tfrac{1}{N}\sum_{n=1}^{N}\mathbf{x}_n$,
which cancels per-point triangulation noise.

\subsection{Phase Detection}
\label{app:phase_detection}

A raw Aria recording interleaves active manipulation with
non-manipulation segments---walking up to the workspace, the
pre-episode scene sweep (App.~\ref{app:triangulation}), and stepping
back once the task ends.
Only the manipulation portions carry clean hand--object dynamics, so
we run an automatic \emph{phase detection} step that segments every
recording into kinematic modes and keeps only the manipulation frames
for training.

\paragraph{Phase taxonomy.}
Each frame is assigned one of five modes:
(0)~\textsc{Manip}---demonstrator stands still and actively
manipulates the scene;
(1)~\textsc{Forward}---linear walking;
(2)~\textsc{Rotate}---in-place head/body rotation (\eg{} the scene
sweep);
(3)~\textsc{Transition}---short buffers between adjacent modes;
(4)~\textsc{Finished}---sustained final hold at the end of the
recording.

\paragraph{Segmentation signals and training-data selection.}
Phases are computed from two streams: the 6-DoF head trajectory from
Aria SLAM (body motion) and the 3D hand trajectory from the hand
tracker (manipulation motion).
A frame enters \textsc{Manip} when the head linear and angular
speeds simultaneously fall below
$v_{\text{stop}}{=}0.03$\,m/s and
$w_{\text{stop}}{=}0.15$\,rad/s for $\ge 15$ consecutive frames;
\textsc{Rotate} requires
$\lVert\boldsymbol{\omega}_{\text{head}}\rVert > 0.10$\,rad/s with
$\lVert\mathbf{v}_{\text{head}}\rVert < 0.08$\,m/s;
\textsc{Forward} collects the remaining high-linear-speed frames;
\textsc{Transition} fills a $10$-frame buffer at every mode change;
and \textsc{Finished} is declared once the trailing stop lasts for
$\ge 30$ frames.
We additionally refine \textsc{Manip} with hand kinematics: a
candidate frame is demoted to \textsc{Transition} if the average
hand speed exceeds $0.15$\,m/s over a $5$-frame window, trimming
reaching/retracting motion away from the manipulation core.
The training pipeline then keeps only \textsc{Manip}~($0$) and
\textsc{Finished}~($4$), dropping \textsc{Forward}, \textsc{Rotate},
and \textsc{Transition}, so the scene sweep, navigation, and
mode-change buffers never reach the training signal.

\subsection{Hand-to-Gripper Transfer}
\label{app:hand2gripper}

\begin{wrapfigure}{r}{0.33\textwidth}
  \vspace{-30pt}
  \centering
  \includegraphics[width=0.31\textwidth]{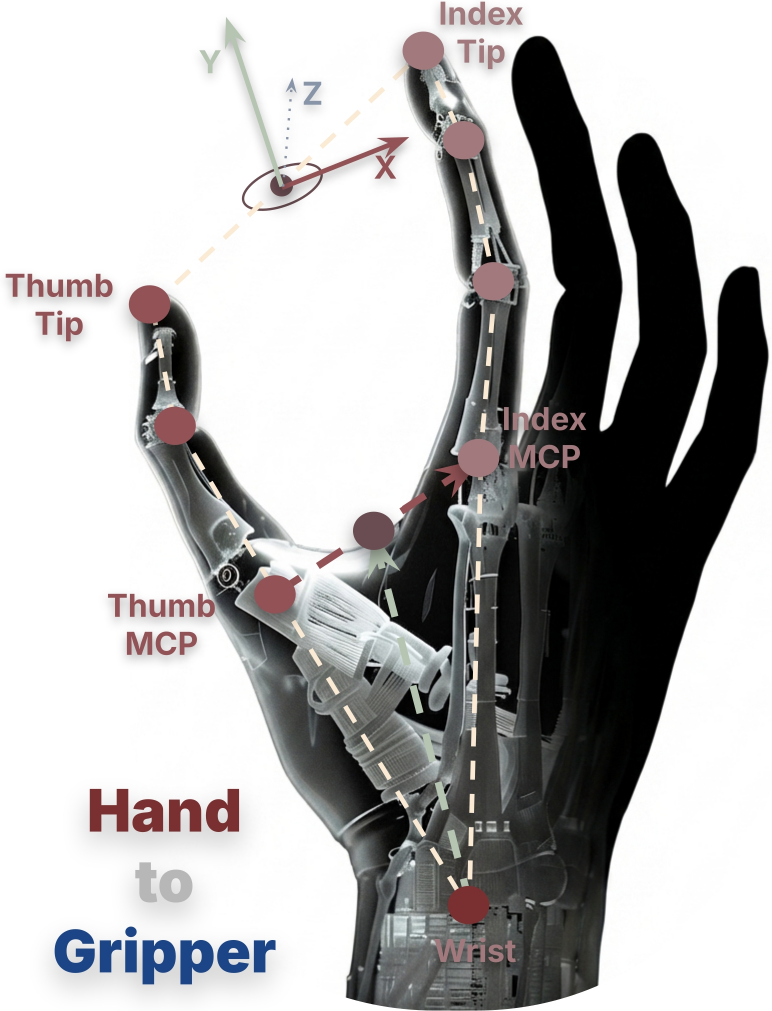}
  \caption{\textbf{Hand-to-gripper mapping.}}
  \label{fig:hand2gripper}
  \vspace{-10pt}
\end{wrapfigure}

To treat a human egocentric video as robot data, every frame of the
demonstration must carry an end-effector target that a parallel-jaw
robot can actually execute.
The human hand, however, has $21$ articulated keypoints and a
morphology very different from a $2$-finger gripper, so the raw hand
pose cannot be passed through directly.
We therefore \emph{retarget} the hand into a virtual gripper---a
6-DoF $\mathrm{SE}(3)$ pose plus a 1-DoF grasp scalar---derived from
a few anatomically stable keypoints after a short motion-optimization
pipeline.

\paragraph{Hand keypoint extraction.}
We start from the $21$-keypoint hand skeleton produced by Aria
MPS~\citep{engel2023aria}, which fuses the stereo SLAM cameras and
the on-device IMU to recover every keypoint's 3D position in the
SLAM world frame at each frame.
For retargeting we use only five keypoints per hand
(Fig.~\ref{fig:hand2gripper}): the \emph{wrist}, \emph{thumb MCP},
\emph{thumb tip}, \emph{index MCP}, and \emph{index tip}.

\paragraph{Motion optimization.}
Raw MPS keypoints are noisy and occasionally drop frames, and
feeding them directly into the $\mathrm{SE}(3)$ construction produces
jittery, flip-prone trajectories.
We therefore run a short optimization pipeline:
(1)~\emph{Confidence masking}---we drop any keypoint whose MPS
confidence falls below $0.8$ and discard detection segments shorter
than $30$ consecutive frames as likely ghost detections;
(2)~\emph{Gap interpolation}---short missing intervals (${\le}10$
frames) are filled with linear interpolation on positions and SLERP
on orientations, so the later smoother sees a dense sequence;
(3)~\emph{Savitzky--Golay position smoothing}---we apply an SG filter
with window size $21$ and polynomial order $2$ to the five retarget
keypoints, removing high-frequency jitter while preserving
manipulation-relevant accelerations;
(4)~\emph{EMA orientation smoothing}---we apply an exponential
moving average with $\alpha_x = \alpha_y = 0.15$ to the $X$- and
$Y$-axes of the gripper frame (defined below), re-orthonormalize via
Gram--Schmidt after each update, and enforce sign consistency across
adjacent frames to prevent spurious $180^{\circ}$ flips.

\paragraph{End-effector position.}
We take the midpoint of the thumb tip and the index tip as the
gripper position, which naturally corresponds to the center of a
parallel-jaw grasp:
\begin{equation}
  \mathbf{p}_{\text{ee}} = \tfrac{1}{2}\big(
    \mathbf{p}_{\text{thumb tip}} + \mathbf{p}_{\text{index tip}}
  \big).
  \label{eq:ee_position}
\end{equation}

\paragraph{End-effector orientation.}
Choosing an orientation that is both \emph{accurate} and
\emph{stable through pinch grasps} is the subtle part of
retargeting; two natural alternatives both fail.
\emph{(i) Raw wrist pose}: using the MPS wrist orientation directly
as the gripper frame is inaccurate, because the anatomical wrist
frame is not aligned with the thumb--index action axis that the
gripper actually uses.
\emph{(ii) Wrist-to-fingertip-midpoint}: defining the forward axis
as $\text{wrist} \to \text{mid}(\text{thumb tip}, \text{index tip})$
and the jaw axis as $\text{thumb tip} \to \text{index tip}$ works
when the hand is open but becomes \emph{degenerate at the moment of
grasp}---the two fingertips converge to nearly the same point, so
the jaw axis collapses to a near-zero vector and the frame is
ill-defined.
We instead build the gripper frame from the MCP joints, which remain
well-separated throughout the full pinch cycle.
Writing $\mathbf{p}_w, \mathbf{p}_{\text{tMCP}}, \mathbf{p}_{\text{iMCP}}$
for the wrist, thumb-MCP, and index-MCP positions, we construct
\begin{equation}
  \mathbf{x}_{\text{ee}} =
    \widehat{\mathbf{p}_{\text{iMCP}} - \mathbf{p}_{\text{tMCP}}},
  \qquad
  \mathbf{y}_{\text{ee}} =
    \widehat{\tilde{\mathbf{y}} -
             (\tilde{\mathbf{y}}^{\top}\mathbf{x}_{\text{ee}})\,\mathbf{x}_{\text{ee}}},
  \qquad
  \mathbf{z}_{\text{ee}} =
    \mathbf{x}_{\text{ee}} \times \mathbf{y}_{\text{ee}},
  \label{eq:ee_orientation}
\end{equation}
where
$\tilde{\mathbf{y}} = \tfrac{1}{2}(\mathbf{p}_{\text{tMCP}} + \mathbf{p}_{\text{iMCP}}) - \mathbf{p}_w$
is the raw wrist-to-MCP-midpoint vector and $\widehat{(\cdot)}$
denotes unit normalization.
Intuitively, $\mathbf{y}_{\text{ee}}$ is the forward axis (wrist
$\rightarrow$ MCP midpoint), $\mathbf{x}_{\text{ee}}$ is the jaw
opening axis (thumb MCP $\rightarrow$ index MCP), and
$\mathbf{z}_{\text{ee}}$ is the orthogonal complement.
The rotation
$R_{\text{ee}} = [\mathbf{x}_{\text{ee}}\,|\,\mathbf{y}_{\text{ee}}\,|\,\mathbf{z}_{\text{ee}}]
\in \mathrm{SO}(3)$ together with the position $\mathbf{p}_{\text{ee}}$
from Eq.~\eqref{eq:ee_position} gives the $\mathrm{SE}(3)$ end-effector
pose $T_{\text{ee}}$.
Because the two MCPs never collapse to each other during a pinch,
this construction stays numerically stable across the full
grasp/release cycle.

\paragraph{Gripper aperture.}
We derive the 1-DoF gripper command from the thumb--index fingertip
distance:
\begin{equation}
  g = \text{clip}\!\left(
    \tfrac{\lVert\mathbf{p}_{\text{thumb tip}} - \mathbf{p}_{\text{index tip}}\rVert - d_{\min}}{d_{\max} - d_{\min}},\;
    0,\; 1
  \right),
  \label{eq:grasp}
\end{equation}
where $d_{\min}$ and $d_{\max}$ are the closed and fully-open
fingertip distances calibrated per user.
The normalized $g$ is then median-filtered and run through a short
flicker-suppression pass to produce a clean open/close command
stream, and binarized at deployment.

%-----------------------------------------------------------------------
\section{Training Details}
\label{app:training_details}

\subsection{Flow Matching Policy}
\label{app:flow_matching_details}

\paragraph{Velocity field and loss.}
We train a conditional flow matching~\citep{lipman2023flow} policy
that maps a Gaussian prior $\mathbf{x}_0 \sim \mathcal{N}(0, I)$
to the ground-truth bimanual action chunk $\mathbf{x}_1 \in
\R^{K \times D_a}$ along the linear path
$\mathbf{x}_t = (1{-}t)\mathbf{x}_0 + t\mathbf{x}_1$ with flow time
$t \sim \mathcal{U}(0,1)$.
The target velocity is the constant displacement
$\mathbf{v}_{\text{target}} = \mathbf{x}_1 - \mathbf{x}_0$, and the
flow-matching loss is an MSE over the predicted velocity with
dimension-wise reweighting:
$w_p{=}5$ on position, $w_r{=}1$ on 6D rotation, and $w_g{=}10$ on
the grasp logit.
We also support an optimal-transport matching variant (OT-CFM) that
solves a Hungarian assignment between noise and action samples within
each mini-batch before computing the loss, producing straighter
target flows; we leave it off by default since we did not find
consistent wins on our tasks.

\paragraph{Network.}
The velocity field $v_\theta$ is a $6$-layer, $8$-head transformer
decoder with embedding dimension $384$ and dropout $0.05$.
Each action-chunk token attends (via self-attention) to the rest of
the chunk and (via cross-attention) to the conditioning context.
Context is built from two streams:
(i)~the RGB frame, embedded with a $16{\times}16$ patch embedding on
a $240{\times}320$ input and a sinusoidal time embedding fused through
a small MLP; and
(ii)~the state tokens, i.e., the per-entity \geo{} tokens described
in Sec.~\ref{sec:spatial}, linearly projected to $384$ channels.

\paragraph{Auxiliary heads.}
Three dense auxiliary objectives share the context encoder with the
velocity field (Sec.~\ref{sec:policy}).
The \emph{object motion} head ($\mathcal{L}_{\text{OM}}$) predicts the
9-D future pose trace of the manipulated object and is trained with
$0.5 (w_p, w_r)$-weighted MSE; the \emph{2D trace} head
($\mathcal{L}_{\text{2D}}$) emits $K{\times}3{\times}2$ normalized image
coordinates of three anchor keypoints through a shallow deconvolution
stack with loss weight $\lambda_{\text{2D}}{=}20$; and the
\emph{latent consistency} head ($\mathcal{L}_{\text{LC}}$) predicts the
hand tokens $K$ steps ahead with a masked MSE weighted by
$\lambda_{\text{LC}}{\in}[0.1, 1.0]$.
All three targets are produced automatically by the perception
pipeline, so each demonstration yields a dense multi-task signal
without extra labeling.

\paragraph{Additional tricks.}
Two lightweight training tricks further stabilize learning from
minutes of data.
\emph{Region attention} biases the image cross-attention toward the
currently active manipulation anchor: given the anchor's 2D image
projection $(u_0, v_0)$, we multiply the attention logits by a
Gaussian spotlight
\begin{equation}
  w(u, v) = \exp\!\Big(-\frac{(u - u_0)^2 + (v - v_0)^2}{2\sigma^2}\Big),
  \label{eq:region_attention}
\end{equation}
whose spatial scale $\sigma$ is a learnable parameter, softly
focusing the encoder on task-relevant image regions without
hard-cropping.
\emph{State-noise injection} perturbs every hand token during
training, $\tilde{s}_t = s_t + \boldsymbol{\epsilon}$ with
$\boldsymbol{\epsilon} \sim \mathcal{N}(\mathbf{0}, \Sigma_s)$ and
separate standard deviations on the position, 6D rotation, and grasp
channels, which makes the policy robust to the small perception
noise it encounters at deployment.

\paragraph{Optimization recipe.}
We train with AdamW at a base learning rate of $10^{-4}$, cosine
decay with $200$-step warmup, minimum-LR ratio $0.05$, batch size
$32$, and $400$ epochs.
We clip gradient norm at $1.0$, use bfloat16 mixed precision, and
keep an exponential moving average of the weights with decay
$0.999$ for evaluation and deployment.

\paragraph{Data augmentation.}
To expand the effective training distribution from only
${\sim}40$\,min of human video per task, we apply a cocktail of
augmentations on the fly in the dataloader, grouped into three
families.
\textit{(i)~Image augmentations} on the RGB stream. Photometric
jitter ($p{=}0.8$) randomly perturbs brightness ($\pm 0.20$),
contrast ($\pm 0.20$), and gamma ($\pm 0.15$), adds Gaussian pixel
noise ($\sigma{=}0.02$), optionally converts the frame to grayscale
($p{=}0.1$), and jitters HSV hue by $\pm 10$ and saturation by
$[0.6, 1.4]$. A random resized crop ($p{=}0.5$) draws a sub-window
with scale in $[0.7, 1.0]$ and aspect ratio in $[0.9, 1.1]$ before
resizing back to the network input size. A Gaussian blur with a
$3{\times}3$ kernel is applied with $p{=}0.15$, and random erasing
($p{=}0.5$) overlays $3$--$8$ black cutout patches each covering
$5$--$20\%$ of the frame area.
\textit{(ii)~Action-target augmentation.} We additively perturb every
target pose in the action chunk with Gaussian noise---
$\sigma_{\text{pos}}{=}1$\,mm on translation and
$\sigma_{\text{rot}}{=}0.5^{\circ}$ on rotation---before the
flow-matching loss is computed, which regularizes the velocity field
against small tracking noise in the labels.
\textit{(iii)~Temporal augmentation.} With $p{=}0.5$ we apply
\emph{sub-step interpolation}: adjacent state/action frames are
linearly blended at a random $\alpha \in [0, 1]$, effectively
densifying the temporal grid at no extra collection cost.

%-----------------------------------------------------------------------
\section{Inference Details}
\label{app:real_world_inference}

\subsection{Deployment Pipeline}
\label{app:deployment}

Nothing inside the policy changes when it moves from human video to a
robot: it consumes the same two inputs as in training---one
embodiment-agnostic RGB image and a set of \geo{} tokens---and only the
modules that \emph{produce} those inputs are swapped.

\textbf{Reference frame.}
At deployment the camera is static, so its frame serves directly as the
episode's reference frame $\mathrm{REF}$: every entity pose and every
predicted action is expressed in it, whichever camera is mounted.
On human video the camera moves with the head, so we instead freeze the
camera pose at the first manipulation frame and use it as
$\mathrm{REF}$. Each demonstration thus contributes its own reference
frame, and a task's demonstrations span the head poses a person
naturally occupies at a table; a new camera placement is another draw
from that distribution rather than an out-of-distribution input.

\textbf{Observation assembly.}
Detection, segmentation, and orientation estimation run \emph{once} per
episode, from one text prompt per object. The robot-side frontend is
the training frontend with triangulation replaced by depth
back-projection, and needs neither SLAM nor the short scene sweep used
to initialize human episodes. Every control cycle we capture an RGB-D
frame, segment and inpaint the robot arm and render the virtual gripper
in its place, and read arm forward kinematics and the gripper state to
obtain $T_{\text{ee}}$ and $g$.
We then assemble each token as in Eq.~\eqref{eq:ict}:
${}^{\mathrm{REF}}\!T_{E}$ in the static camera frame,
${}^{E}\!T_{LH}$ and ${}^{E}\!T_{RH}$ from the gripper pose, and $g$
from the gripper opening.
Latching carries over unchanged from training (Sec.~\ref{sec:spatial}),
with the end effector in place of the hand: from grasp onset the object
pose is propagated by the gripper rather than re-estimated, assuming no
in-hand manipulation: a grasped object keeps a fixed pose relative to
the gripper. This keeps the object token correct under
occlusion, but a physical slip out of the gripper still ends the
episode in failure.

\textbf{Control loop.}
The policy integrates the learned flow field into a 50-step action
chunk, decodes camera-frame end-effector targets, and maps them to arm commands through the
hand--eye extrinsic. The loop is synchronous at ${\sim}10$\,Hz on a
single RTX~4090, so each plan acts on a frame at most ${\sim}100$\,ms
old while the predicted action chunk executes continuously between
plans. Token assembly is negligible; runtime is dominated by arm
inpainting and segmentation.

\textbf{Requirements.}
The only per-rig setup is camera intrinsics and one hand--eye extrinsic
per arm, both obtained once and reused across tasks. We use no CAD
models, no manual initialization, no online object tracking, and no
depth inside the policy---depth serves the frontend only.
Hardware-agnostic therefore means no robot demonstrations and no
task-specific calibration: the same checkpoint drives a different arm
or camera once its extrinsic is known.

\subsection{Robot Inference Setup}
\label{app:robot_inference_setup}

\begin{wrapfigure}{r}{0.48\textwidth}
    \vspace{-30pt}
    \centering
    \includegraphics[width=0.46\textwidth]{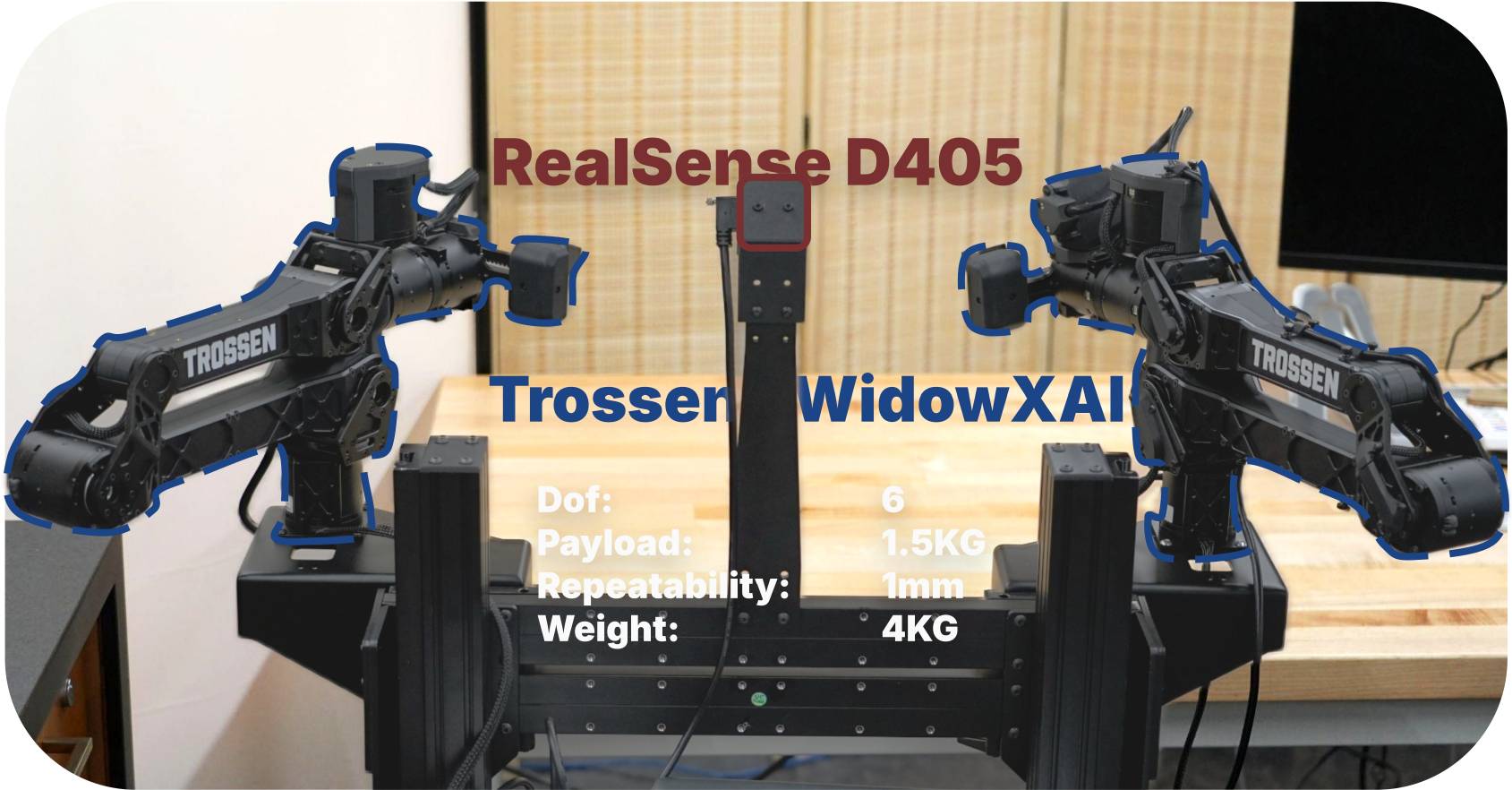}
    \caption{\textbf{Robot inference setup.}}
    \label{fig:inference_setup}
    \vspace{-10pt}
\end{wrapfigure}

Apart from the zero-shot generalization study
(Sec.~\ref{sec:generalization}), all real-world experiments in the
main paper use the single inference setup shown in
Fig.~\ref{fig:inference_setup}: two Trossen WidowX AI arms mounted
side-by-side on a shared workbench, forming a bimanual platform
that handles both single-arm and two-arm tasks without any hardware
change between tasks.
Each WidowX AI arm is a 6-DoF parallel-jaw manipulator with a
${\sim}1.5$\,kg payload at full reach and $\pm 1$\,mm end-effector
repeatability.
For visual input we use a single Intel RealSense D405 mounted
top-down above the workspace; its RGB stream is the sole observation
consumed by \ours{}.
Each WidowX AI arm also ships with a built-in wrist camera, but we
deliberately do not use it for \ours{}: the robot-teleoperation ACT
baseline~\citep{zhao2023act} in Sec.~\ref{sec:main_results}, in
contrast, does consume the wrist cameras as part of its standard
observation interface.

\subsection{Flow Matching Rollout and Control}
\label{app:inference_details}

\paragraph{ODE rollout.}
At test time we integrate the learned velocity field with a
fixed-step Euler solver using $20$ inference steps: starting from a
noise sample $\mathbf{x}_0 \sim \mathcal{N}(0, I)$ drawn once at
policy load time, we iterate $\mathbf{x}_{t+\Delta t} \leftarrow
\mathbf{x}_t + v_\theta(\mathbf{x}_t, t, s_t)\,\Delta t$ with
$\Delta t = 1/20$, yielding a $K{=}50$-step bimanual action chunk in
one forward pass per re-plan.
Predictions are unpacked dimension-wise into per-hand position,
6D rotation, and grasp logit, with positions denormalized by the
dataset mean/std, rotations projected back to $\mathrm{SO}(3)$ via
\emph{normalize-then-Gram--Schmidt} on the 6D representation, and
grasps passed through a sigmoid.

\paragraph{Action chunking and control.}
The controller re-plans at every cycle ($10$\,Hz), keeping at most
one prediction in history and executing one action per cycle.
A step stride of $2$ sub-samples the chunk so the effective executed
rate is $5$\,Hz, and a look-ahead offset of $25$ steps lets the
controller query the chunk slightly ahead of the current execution
index to mask planning latency.
For grasp we use an \texttt{any}-over-horizon rule: the gripper
closes as soon as \emph{any} step in the current chunk predicts a
grasp probability above $0.6$; an optional grasp-latch mode
additionally locks the gripper closed after the first grasp event
to prevent accidental mid-task releases.

\paragraph{Smoothing and safety.}
To hide small noise in the predicted SE(3) stream we apply an EMA on
positions ($\alpha{=}0.5$) and quaternion SLERP on rotations before
streaming targets to the arms, and a trajectory-overlap blend
(smoothing parameter $12$) to avoid jerky starts/stops between
consecutive chunks.
Finally, a safety cage limits each per-cycle target displacement to
${\le}0.08$\,m in position and ${\le}0.02$\,rad in rotation to
guard against sudden outliers; we did not observe any safety-cage
clamp during normal rollouts in our experiments.

%-----------------------------------------------------------------------
\section{Additional Experiments Analysis}
\label{app:analysis}

\subsection{Hand Tracking Method Study}
\label{app:hand_tracking}

\begin{figure}[h]
    \centering
    \includegraphics[width=\linewidth]{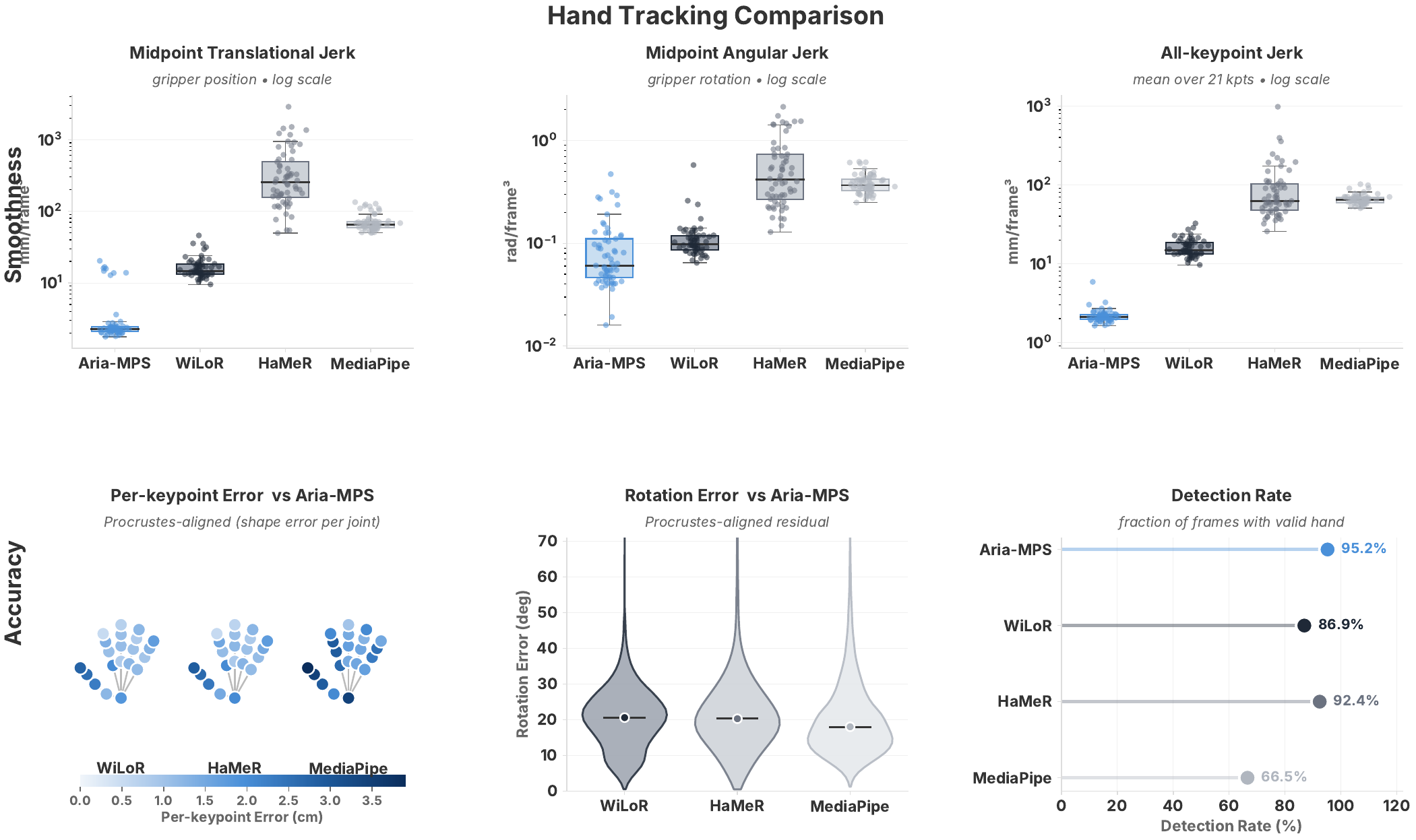}
    \caption{\textbf{Hand tracking comparison on Serve Bread} (45
    demonstrations, ${\sim}45$\,k frames).
    \emph{Top---Smoothness:} per-frame jerk of the gripper midpoint
    (translational and angular) and of all 21 keypoints
    (lower is better, log scale).
    \emph{Bottom---Accuracy vs.\ Aria-MPS:} per-keypoint shape error
    after Procrustes alignment, residual rotation error after
    subtracting the systematic frame offset, and fraction of frames
    with a valid hand detection.}
    \label{fig:hand_tracking}
\end{figure}

\paragraph{Setup.}
\geo{} consumes 3D hand keypoints as input, so the quality of the
upstream hand tracker directly affects what the policy can learn.
We isolate this dependency on Serve Bread by holding everything else
constant---the same 45 demonstrations (30\,min total), the same
\ours{} architecture, the same training recipe---and varying only the
hand-tracking module that produces the action labels.
We compare four trackers spanning the dominant design choices in the
literature:
(1) \textbf{Aria-MPS}~\citep{engel2023aria}, our default, which fuses
the two wide-FoV \emph{monochrome SLAM} cameras with the on-device
IMU through Meta's MPS pipeline to recover metric 3D keypoints---note
that the central RGB camera is used only for video logging, not for
hand tracking;
(2) \textbf{WiLoR}~\citep{potamias2024wilor}, a transformer that
regresses MANO parameters from a single RGB crop per frame;
(3) \textbf{HaMeR}~\citep{pavlakos2024hamer}, a strong monocular RGB
estimator that also predicts MANO parameters but processes frames
independently---temporal and world-frame extensions of this family
have since been proposed~\citep{yu2025dynhamr,zhang2025hawor};
and (4) \textbf{MediaPipe}~\citep{lugaresi2019mediapipe}, a lightweight
monocular RGB pipeline whose 3D outputs are root-relative and have to
be lifted with the camera depth.
\begin{wrapfigure}{r}{0.48\textwidth}
    \vspace{-14pt}
    \centering
    \includegraphics[width=0.46\textwidth]{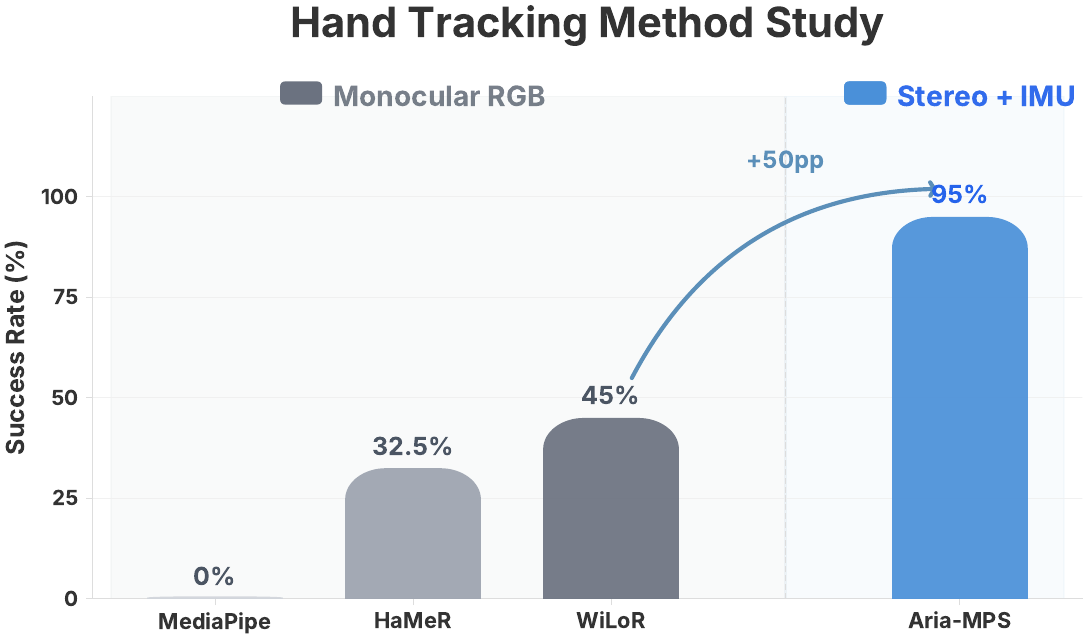}
    \caption{\textbf{Hand Tracking Method Study.}}
    \label{fig:hand_tracking_results}
    \vspace{-10pt}
\end{wrapfigure}

Beyond these vision-only trackers, instrumented alternatives recover
hand pose from multi-modal sensing gloves~\citep{zhang2026glove2hand}
or 6-axis IMU systems~\citep{sarker2026imuhand}, but require
specialized hardware on the demonstrator and fall outside our
zero-instrumentation collection setup.
For each tracker we re-run the data preprocessing, train
\ours{} from scratch, and evaluate 40 real-world trials on Serve Bread
(Fig.~\ref{fig:hand_tracking_results}).
Aria-MPS is treated as the reference for all alignment-style metrics
in Fig.~\ref{fig:hand_tracking}.

\paragraph{Results.}
Fig.~\ref{fig:hand_tracking_results} shows real-world success
collapsing by $95 \to 45 \to 32.5 \to 0$\,\% as we move from stereo
Aria-MPS to monocular trackers, and Fig.~\ref{fig:hand_tracking}
explains why.
We highlight four observations.

\textbf{Stereo depth is decisive for downstream success.}
Real-world success drops from 95\,\% (Aria-MPS) to at most 45\,\%
(WiLoR) the moment we replace stereo with monocular RGB.
Monocular networks are inherently scale-ambiguous along the depth
axis, producing a 5--11\,cm systematic depth offset that propagates
directly into the \geo{} reference frame, so the policy never learns
a consistent grasp.

\textbf{Smoothness and tracking persistence---not pose accuracy
per se---separate the surviving baselines.}
After Procrustes alignment the per-keypoint residual error is nearly
identical for HaMeR and WiLoR (1.4\,cm vs.\ 1.4\,cm; both dominated
by the shared MANO inductive bias) and only modestly worse for
MediaPipe (2.2\,cm).
Yet WiLoR (45\,\%) clearly beats HaMeR (32.5\,\%).
The gap is explained by the smoothness panels: WiLoR's
gripper-midpoint jerk is more than an order of magnitude lower than
HaMeR's---its per-frame predictions happen to be far more
temporally stable---and its detection rate is 86.9\,\% vs.\
MediaPipe's 66.5\,\%.
A jittery or intermittently missing trajectory teaches the policy
incoherent action labels.

\textbf{The MANO prior is a double-edged sword.}
HaMeR and WiLoR show essentially identical per-keypoint Procrustes
residuals (Fig.~\ref{fig:hand_tracking}, bottom-left): the colored
patterns on the two hand skeletons are visually indistinguishable.
This is not a coincidence---both networks regress MANO pose
parameters and inherit the same canonical bone proportions.
The shared shape prior means their residual shape errors are
correlated by construction, while methods without a parametric
constraint (Aria-MPS, MediaPipe) deviate in different ways.
This explains why pose accuracy alone is a poor predictor of
downstream policy success.

\textbf{MediaPipe fails entirely.}
With 0\,\% real-world success and only 66.5\,\% detection rate,
MediaPipe cannot produce coherent action labels even with the
\geo{} representation absorbing some noise.
Its 3D output is root-relative and must be lifted with the camera
depth, and the lift collapses the hand thickness to ${\sim}7$\,cm
(vs.\ ${\sim}16$\,cm for Aria), flattening the pose information
that \geo{} relies on.

Together these results point to a clear practical message: invest
in the perception frontend.
A more accurate hand tracker---especially one that exploits stereo
or learned depth---is the highest-leverage upgrade for any policy
that operates on hand-derived spatial tokens.

\subsection{Human-Robot Co-Training Study}
\label{app:cotraining}

\paragraph{Setup.}
The data-efficiency study (Sec.~\ref{sec:data_efficiency}) showed
that human demonstrations are roughly $3.75\times$ more sample-efficient
than robot teleoperation \emph{at the same total collection time}.
Here we ask the complementary question: when both modalities are
available, what is the best mixing ratio?
We hold the total collection time fixed at 30\,min and vary the
fraction of \emph{human} data from 0\,\% (pure robot teleoperation,
${\sim}45$ teleop episodes) to 100\,\% (pure human egocentric video,
${\sim}45$ egocentric demos) in 25-pp steps.
Each batch is sampled with the corresponding ratio, so the policy
sees both modalities in the intended proportion at every gradient
step.
The architecture, optimizer, schedule, and number of training steps
are kept identical across the five conditions; only the data mixture
changes.
We evaluate 40 real-world trials on Serve Bread per condition.

\paragraph{Results.}
\begin{wrapfigure}{r}{0.48\textwidth}
    \vspace{-16pt}
    \centering
    \includegraphics[width=0.46\textwidth]{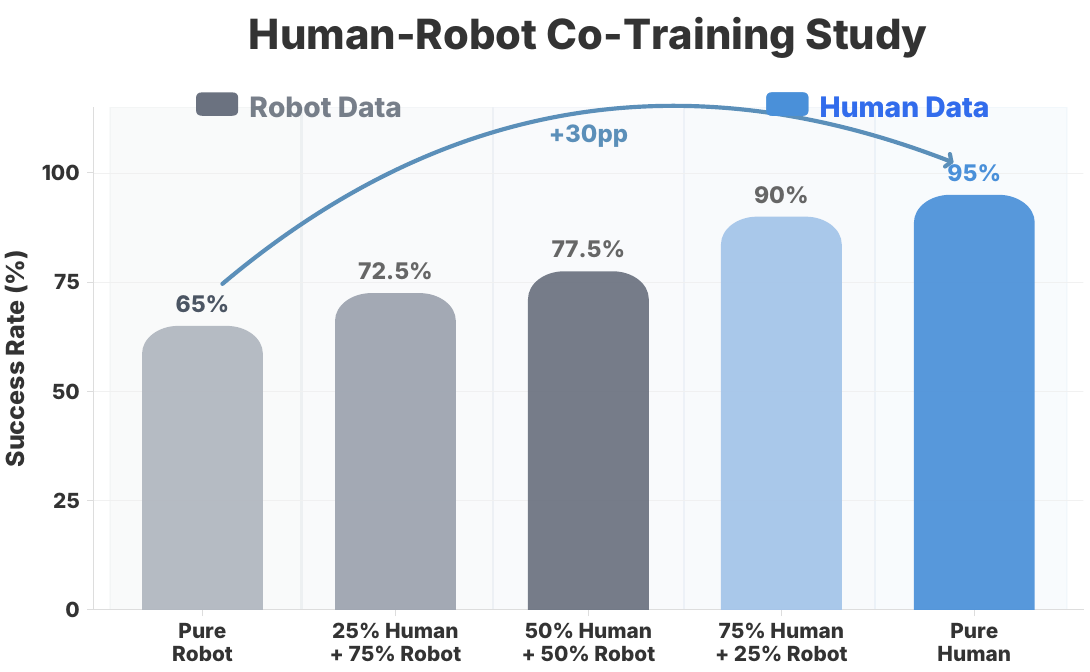}
    \caption{\textbf{Human-Robot Co-Training Study.}}
    \label{fig:cotraining}
    \vspace{-10pt}
\end{wrapfigure}
Real-world success increases \emph{monotonically} as the human-data
ratio grows: $65 \to 72.5 \to 77.5 \to 90 \to 95$\,\% for human
ratios of $0/25/50/75/100$\,\% (Fig.~\ref{fig:cotraining}).
The pure-human policy improves over the pure-robot policy by
$+30$\,pp---a gap larger than that between most of our baselines.
We extract two main findings.

\textbf{Even a small slice of human data dominates.}
Replacing just 25\,\% of the robot teleop with egocentric video
already lifts success from 65\,\% to 72.5\,\% ($+7.5$\,pp),
even though the absolute amount of robot data drops from 30\,min
to 22.5\,min in that condition.
A naive ``more data is always better'' view would predict the
opposite: less robot data should hurt.
Instead the policy improves, indicating that the marginal robot
minutes contribute less signal than the marginal human minutes.
This reproduces the data-efficiency conclusion
(Sec.~\ref{sec:data_efficiency}) at the level of the gradient
step: the policy is happier learning from a small amount of clean
egocentric video than from a large amount of teleop trajectory.

\textbf{The more human data, the better---no co-training sweet spot.}
Across all four transitions the curve only goes up, and the pure
human policy is the global maximum.
We find no ``sweet spot'' where mixing in robot data outperforms
the human-only condition---in fact 75/25 (90\,\%) is already
$5$\,pp below 100/0 (95\,\%), so adding even 7.5\,min of robot
teleoperation actively erodes a 22.5\,min human dataset.
We attribute this to the higher per-minute information density of
human demonstrations documented in Sec.~\ref{sec:data_efficiency}
(Fig.~\ref{fig:data_comparison}): human videos exhibit higher
signal-to-noise ratio, an order-of-magnitude smoother trajectories,
near-zero idle time, and broader spatial coverage than robot
teleoperation.
At a fixed compute and time budget, the policy is best served by
spending every minute on human data.
Combined with the embodiment-invariance of \geo{}, this means a
practitioner deploying \ours{} should not invest in robot
teleoperation at all---the same budget collected as egocentric
human video yields a strictly better policy.

\section{Hyperparameters}
\label{app:hyperparameters}

Table~\ref{tab:hyperparameters_full} consolidates every hyperparameter
value that appears in the main paper and the appendix, grouped by
pipeline stage. Unless explicitly noted, a single value is used
across all four tasks, all training runs, and all real-world trials.

{\small
\begin{longtable}{@{}l r@{}}
\caption{\textbf{All hyperparameters used in \ours{}.}
Values are shared across the four tasks unless noted.}
\label{tab:hyperparameters_full} \\
\toprule
\textbf{Parameter} & \textbf{Value} \\
\midrule
\endfirsthead
\multicolumn{2}{c}{\small\textit{Table~\ref{tab:hyperparameters_full} -- continued from previous page}} \\
\toprule
\textbf{Parameter} & \textbf{Value} \\
\midrule
\endhead
\midrule
\multicolumn{2}{r@{}}{\small\textit{Continued on next page}} \\
\endfoot
\bottomrule
\endlastfoot
\multicolumn{2}{@{}l}{\textit{Data Collection (App.~\ref{app:aria_setup}, App.~\ref{app:task_details})}} \\
Demonstrations per task                           & $45$ \\
Total human-video time per task                   & $30$\,min \\
RGB stream rate / resolution                      & $30$\,fps $/$ $2$\,MP \\
SLAM cameras: count, rate, resolution             & $2$, $30$\,fps, VGA \\
Eye-tracking cameras: count, rate, resolution     & $2$, $10$\,fps, QVGA \\
IMU rates                                         & $1000$\,Hz, $800$\,Hz \\
Pre-episode scene-sweep duration                  & ${\sim}1$\,s \\
Scene-sweep frame count                           & ${\sim}30$ \\
\midrule
\multicolumn{2}{@{}l}{\textit{Phase Detection (App.~\ref{app:phase_detection})}} \\
Head linear-speed stop threshold $v_{\text{stop}}$ & $0.03$\,m/s \\
Head angular-speed stop threshold $w_{\text{stop}}$ & $0.15$\,rad/s \\
Minimum stop-hold duration                        & $15$\,frames \\
Rotate trigger $w_{\text{rot}}$                   & $0.10$\,rad/s \\
Rotate max linear speed $v_{\text{rot,max}}$      & $0.08$\,m/s \\
Transition buffer                                 & $10$\,frames \\
Hand-velocity demotion threshold $v_{\text{hand}}$ & $0.15$\,m/s \\
Hand-velocity averaging window                    & $5$\,frames \\
Finished-stop length                              & $30$\,frames \\
\midrule
\multicolumn{2}{@{}l}{\textit{Hand-to-Gripper Retargeting (App.~\ref{app:hand2gripper})}} \\
Retarget keypoints per hand                       & $5$ \\
MPS confidence threshold                          & $0.8$ \\
Minimum detection segment                         & $30$\,frames \\
Max gap for interpolation                         & $10$\,frames \\
Savitzky--Golay window                            & $21$\,frames \\
Savitzky--Golay polynomial order                  & $2$ \\
EMA smoothing factor $\alpha_x = \alpha_y$        & $0.15$ \\
\midrule
\multicolumn{2}{@{}l}{\textit{Policy Network (App.~\ref{app:flow_matching_details})}} \\
Transformer layers                                & $6$ \\
Transformer attention heads                       & $8$ \\
Transformer embedding dim                         & $384$ \\
Dropout                                           & $0.05$ \\
RGB patch size                                    & $16 \times 16$ \\
RGB input resolution                              & $240 \times 320$ \\
\geo{} token dim                                  & $29$ \\
Prediction horizon $K$                            & $50$ \\
\midrule
\multicolumn{2}{@{}l}{\textit{Losses (App.~\ref{app:flow_matching_details})}} \\
Position weight $w_p$                             & $5$ \\
Rotation weight $w_r$                             & $1$ \\
Grasp weight $w_g$                                & $10$ \\
Object-motion weight $\lambda_{\text{OM}}$ (pos / rot) & $0.5\,w_p$ / $0.5\,w_r$ \\
2D-trace weight $\lambda_{\text{2D}}$             & $20$ \\
Latent-consistency weight $\lambda_{\text{LC}}$   & $[0.1, 1.0]$ \\
\midrule
\multicolumn{2}{@{}l}{\textit{Optimization (App.~\ref{app:flow_matching_details})}} \\
Optimizer                                         & AdamW \\
Base learning rate                                & $1\times 10^{-4}$ \\
Warmup steps                                      & $200$ \\
Min-LR ratio                                      & $0.05$ \\
Batch size                                        & $32$ \\
Epochs                                            & $400$ \\
Gradient-norm clip                                & $1.0$ \\
EMA decay                                         & $0.999$ \\
\midrule
\multicolumn{2}{@{}l}{\textit{Data Augmentation (App.~\ref{app:flow_matching_details})}} \\
Photometric jitter probability                    & $0.8$ \\
\quad Brightness / contrast delta                 & $\pm 0.20$ / $\pm 0.20$ \\
\quad Gamma delta                                 & $\pm 0.15$ \\
\quad Pixel noise $\sigma$                        & $0.02$ \\
\quad Grayscale probability                       & $0.1$ \\
\quad HSV hue jitter                              & $\pm 10$ \\
\quad HSV saturation range                        & $[0.6, 1.4]$ \\
Random resized crop probability                   & $0.5$ \\
\quad Scale range                                 & $[0.7, 1.0]$ \\
\quad Aspect-ratio range                          & $[0.9, 1.1]$ \\
Gaussian blur probability                         & $0.15$ \\
\quad Kernel size                                 & $3 \times 3$ \\
Random erasing probability                        & $0.5$ \\
\quad Number of holes                             & $3$--$8$ \\
\quad Per-hole area                               & $5$--$20\%$ \\
Target position noise $\sigma_{\text{pos}}$       & $1$\,mm \\
Target rotation noise $\sigma_{\text{rot}}$       & $0.5^{\circ}$ \\
Sub-step interpolation probability                & $0.5$ \\
\midrule
\multicolumn{2}{@{}l}{\textit{Hardware (App.~\ref{app:robot_inference_setup})}} \\
Robot arms                                        & $2 \times$ Trossen WidowX AI \\
WidowX AI DoF                                     & $6$ \\
WidowX AI payload at full reach                   & ${\sim}1.5$\,kg \\
WidowX AI end-effector repeatability              & $\pm 1$\,mm \\
Inference RGB camera                              & Intel RealSense D405 (top-mounted) \\
\midrule
\multicolumn{2}{@{}l}{\textit{Inference (App.~\ref{app:inference_details})}} \\
Euler ODE steps                                   & $20$ \\
Step size $\Delta t$                              & $1/20$ \\
Executed action-chunk length                      & $K = 50$ \\
Control-loop frequency                            & $10$\,Hz \\
Action-step stride                                & $2$ \\
Effective executed rate                           & $5$\,Hz \\
Look-ahead offset                                 & $25$\,steps \\
Grasp probability threshold                       & $0.6$ \\
Position smoothing EMA $\alpha$                   & $0.5$ \\
Rotation smoothing                                & quaternion SLERP \\
Trajectory-overlap smoothing parameter            & $12$ \\
Safety cage: max position step                    & $0.08$\,m \\
Safety cage: max rotation step                    & $0.02$\,rad \\
\midrule
\multicolumn{2}{@{}l}{\textit{Robot Teleoperation for ACT Baseline}} \\
Teleop control frequency (leader $\to$ follower)  & $200$\,Hz \\
Teleop recording frequency                        & $30$\,Hz \\
Leader $\to$ follower EMA $\alpha$                & $0.5$ \\
Top camera resolution                             & $640 \times 480$ \\
Wrist camera resolution                           & $320 \times 240$ \\
Action space                                      & $7$-DoF joint positions \\
Proprioception dim                                & $7$ \\
\midrule
\multicolumn{2}{@{}l}{\textit{ACT Baseline Training}} \\
Visual backbone                                   & ResNet-18 (pretrained) \\
Embedding dim                                     & $256$ \\
Attention heads                                   & $8$ \\
Encoder / decoder layers                          & $4$ / $1$ \\
Feed-forward dim                                  & $2048$ \\
CVAE latent dim                                   & $32$ \\
Dropout                                           & $0.1$ \\
Prediction horizon $K$                            & $50$ \\
Image input resolution                            & $240 \times 320$ \\
Batch size                                        & $24$ \\
Epochs                                            & $400$ \\
Base learning rate                                & $1 \times 10^{-4}$ \\
Weight decay                                      & $1 \times 10^{-2}$ \\
Warmup steps                                      & $500$ \\
EMA decay                                         & $0.999$ \\
Action L1 weight $w_{\text{pos}}$                 & $5.0$ \\
CVAE KL loss weight (annealed to)                 & $10.0$ \\
\end{longtable}
}

\end{document}